\documentclass{article} 
\usepackage{times}
\usepackage{graphicx} 

\usepackage{natbib}

\usepackage{algorithm}
\usepackage{algorithmic}

\usepackage{hyperref}

\usepackage[accepted]{icml2017}

\usepackage{wrapfig}
\usepackage{enumitem}
\usepackage{subcaption}
\usepackage[font={small}]{caption}
\usepackage{amsmath}
\usepackage{amsthm}
\usepackage{amssymb}
\usepackage{multirow}
\usepackage{pifont}

\icmltitlerunning{Zero-Shot Task Generalization with Multi-Task Deep Reinforcement Learning}

\begin{document} 

\twocolumn[
\icmltitle{Zero-Shot Task Generalization with Multi-Task Deep Reinforcement Learning}


\icmlsetsymbol{equal}{*}

\begin{icmlauthorlist}
\icmlauthor{Junhyuk Oh}{umich}
\icmlauthor{Satinder Singh}{umich}
\icmlauthor{Honglak Lee}{umich,google}
\icmlauthor{Pushmeet Kohli}{msr}
\end{icmlauthorlist}

\icmlaffiliation{umich}{University of Michigan}
\icmlaffiliation{msr}{Microsoft Research}
\icmlaffiliation{google}{Google Brain}

\icmlcorrespondingauthor{Junhyuk Oh}{junhyuk@umich.edu}


\vskip 0.3in
]
\printAffiliationsAndNotice{}

\newcommand{\hit}{transform}
\newcommand{\hitting}{transforming}
\newcommand{\Hit}{Transform}
\newcommand{\Hitting}{Transforming}

\newcommand{\multi}{subtask controller}
\newcommand{\Multi}{Subtask controller}
\newcommand{\supplementary}{Appendix}
\newcommand\REMOVE[1]{\textcolor{red}{#1}}

\iffalse 
        \newcommand{\todo}[1]{}
        \newcommand{\outline}[1]{}
        \newcommand{\textgray}[1]{}
        \newcommand{\commenttext}[1]{}
        \newcommand{\commentfoot}[1]{}
        \newcommand{\commentselfoot}[2]{}
        \newcommand{\commentselrep}[2]{}
        \newcommand{\topic}[1]{}
\else 
        \newcommand{\todo}[1]{{\textcolor{red}{[[TODO: {#1}]]}}}
        \newcommand{\outline}[1]{{\textcolor{blue}{[[{#1}]]}}}
        \newcommand{\textgray}[1]{\textcolor{gray}{[[{#1}]]}}
        \newcommand{\commenttext}[1]{\textcolor{red}{[[{#1}]]}}
        \newcommand{\commentfoot}[1]{\footnote{\textcolor{red}{\textit{#1}}}}
        \newcommand{\commentselfoot}[2]{{\textcolor{blue}{#1}}\commenttext{#2}}
        \newcommand{\commentselrep}[2] {{\textcolor{blue}{#1}} {\textcolor{green}{[[\textit{#2}]]}}}
        \newcommand{\topic}[1]{\textcolor{gray}{\textbf{(#1.)}}}
\fi

\newcommand{\cutabstractup}{\vspace*{-0.2in}}
\newcommand{\cutabstractdown}{\vspace*{-0.00in}}

\newcommand{\cutsectionup}{\vspace*{-0.05in}}
\newcommand{\cutsectiondown}{\vspace*{-0.05in}}

\newcommand{\cutsubsectionup}{\vspace*{-0.05in}}
\newcommand{\cutsubsectiondown}{\vspace*{-0.05in}}

\newcommand{\cutsubsubsectionup}{\vspace*{-0.02in}}
\newcommand{\cutsubsubsectiondown}{\vspace*{-0.02in}}

\newcommand{\cutcaptionup}{\vspace*{-0.0in}}
\newcommand{\cutcaptiondown}{\vspace*{-0.0in}}

\newcommand{\cutparagraphup}{\vspace*{-0.05in}}
\newcommand{\cutparagraphdown}{\vspace*{-0.02in}}


\cutabstractup
\begin{abstract}
As a step towards developing zero-shot task generalization capabilities in reinforcement learning (RL), we introduce a new RL problem where the agent should learn to execute sequences of instructions after learning useful skills that solve subtasks. In this problem, we consider two types of generalizations: to previously unseen instructions and to longer sequences of instructions. For generalization over unseen instructions, we propose a new objective which encourages learning correspondences between similar subtasks by making analogies. For generalization over sequential instructions, we present a hierarchical architecture where a meta controller learns to use the acquired skills for executing the instructions. To deal with delayed reward, we propose a new neural architecture in the meta controller that learns when to update the subtask, which makes learning more efficient.
Experimental results on a stochastic 3D domain show that the proposed ideas are crucial for generalization to longer instructions as well as unseen instructions.
\vspace*{-0.2in}
\end{abstract}

\cutsectionup
\section{Introduction} \label{sec:Introduction}
\cutsectiondown 

The ability to understand and follow instructions allows us to perform a large number of new complex sequential tasks even without additional learning.
For example, we can make a new dish following a recipe, and explore a new city following a guidebook. 
Developing the ability to execute instructions can potentially allow reinforcement learning (RL) agents to generalize quickly over tasks for which such instructions are available.
For example, factory-trained household robots could execute novel tasks in a new house following a human user's instructions (e.g., tasks involving household chores, going to a new place, picking up/manipulating new objects, etc.). 
In addition to generalization over instructions, an intelligent agent should also be able to handle unexpected events (e.g., low battery, arrivals of reward sources) while executing instructions. Thus, the agent should not blindly execute instructions sequentially but sometimes deviate from instructions depending on circumstances, which requires balancing between two different objectives.

\begin{figure}
\centering
		\vspace{-5pt}
\includegraphics[width=0.90\linewidth]{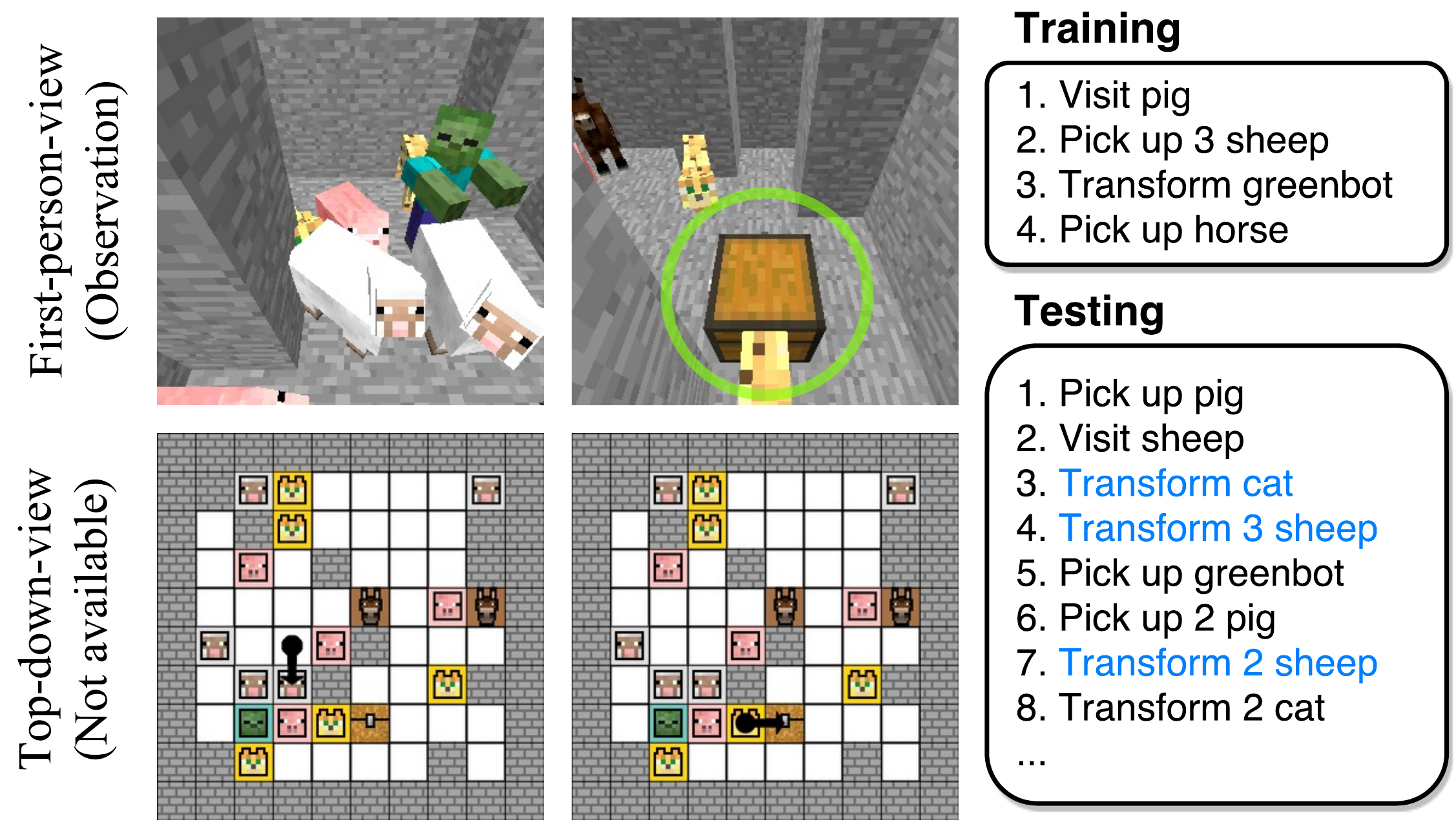} 
		\vspace{-5pt}
   	    \caption{Example of 3D world and instructions. 
   	    The agent is tasked to execute longer sequences of instructions in the correct order after training on short sequences of instructions; in addition, previously unseen instructions can be given during evaluation (blue text). Additional reward is available from randomly appearing boxes regardless of instructions (green circle). }
   	    \vspace{-10pt}
   	    \label{fig:problem}
\end{figure}

\cutparagraphup
\paragraph{Problem.}
To develop such capabilities, this paper introduces the instruction execution problem where the agent's overall task is to execute a given list of instructions described by a simple form of natural language while dealing with unexpected events, as illustrated in Figure~\ref{fig:problem}. More specifically, we assume that each instruction can be executed by performing one or more high-level subtasks in sequence.
Even though the agent can pre-learn skills to perform such subtasks (e.g., [Pick up, Pig] in Figure~\ref{fig:problem}), and the instructions can be easily translated to subtasks, our problem is difficult due to the following challenges. 
\vspace{-10pt}
\begin{itemize}[leftmargin=*]
\setlength\itemsep{0em}
\item \emph{Generalization}: Pre-training of skills can only be done on a subset of subtasks, but the agent is required to perform previously unseen subtasks (e.g., going to a new place) in order to execute unseen instructions during testing. Thus, the agent should learn to generalize to new subtasks in the skill learning stage. Furthermore, the agent is required to execute previously unseen and longer sequences of instructions during evaluation. 
\item \emph{Delayed reward}: The agent is \emph{not} told which instruction to execute at any point of time from the environment but just given the full list of instructions as input. In addition, the agent does \emph{not} receive any signal on completing individual instructions from the environment, i.e., success-reward is provided only when all instructions are executed correctly. Therefore, the agent should keep track of which instruction it is executing and decide when to move on to the next instruction. 
\item \emph{Interruption}: As described in Figure~\ref{fig:problem}, there can be unexpected events in uncertain environments, such as opportunities to earn bonuses (e.g., windfalls), or emergencies (e.g., low battery). It can be better for the agent to \textit{interrupt} the ongoing subtask before it is finished, perform a different subtask to deal with such events, and resume executing the interrupted subtask in the instructions after that. Thus, the agent should achieve a balance between executing instructions and dealing with such events. 
\item \emph{Memory}: There are loop instructions (e.g., ``Pick up 3 pig'') which require the agent to perform the same subtask ([Pick up, Pig]) multiple times and take into account the history of observations and subtasks in order to decide when to move on to the next instruction correctly. 
\end{itemize}
\vspace{-8pt}
Due to these challenges, the agent should be able to execute a novel subtask, keep track of what it has done, monitor observations to interrupt ongoing subtasks depending on circumstances, and switch to the next instruction precisely when the current instruction is finished.

\cutparagraphup
\paragraph{Our Approach and Technical Contributions.}
To address the aforementioned challenges, we divide the learning problem into two stages: 1) learning skills to perform a set of subtasks and generalizing to unseen subtasks, and 2) learning to execute instructions using the learned skills. 
Specifically, we assume that subtasks are 
defined by several disentangled parameters. Thus, in the first stage our architecture learns a \textit{parameterized skill}~\citep{Silva2012LearningPS} to perform different subtasks depending on input parameters. In order to generalize over unseen parameters, we propose a new objective function that encourages making analogies between similar subtasks so that the underlying manifold of the entire subtask space can be learned without experiencing all subtasks. 
In the second stage, our architecture learns a meta controller on top of the parameterized skill so that it can read instructions and decide which subtask to perform. The overall hierarchical RL architecture is shown in Figure~\ref{fig:overview}. To deal with delayed reward as well as interruption, we propose a novel neural network (see Figure~\ref{fig:arch-meta}) that learns when to update the subtask in the meta controller. This not only allows learning to be more efficient under delayed reward by operating at a larger time-scale but also allows interruptions of ongoing subtasks when an unexpected event is observed.

\cutparagraphup
\paragraph{Main Results.}
We developed a 3D visual environment using Minecraft based on~\citet{oh2016memory} where the agent can interact with many objects. Our results on multiple sets of parameterized subtasks show that our proposed analogy-making objective can generalize successfully. 
Our results on multiple instruction execution problems show that our meta controller's ability to learn when to update the subtask plays a key role in solving the overall problem and outperforms several hierarchical baselines. 
The demo videos are available at the following website: \url{https://sites.google.com/a/umich.edu/junhyuk-oh/task-generalization}.

The rest of the sections are organized as follows. Section~\ref{sec:related-work} presents related work. Section~\ref{sec:multi-task} presents our analogy-making objective for generalization to parameterized tasks and demonstrates its application to different generalization scenarios. Section~\ref{sec:meta} presents our hierarchical architecture for the instruction execution problem with our new neural network that learns to operate at a large time-scale. In addition, we demonstrate our agent's ability to generalize over sequences of instructions, as well as provide a comparison to several alternative approaches.


\cutsectionup
\section{Related Work} \label{sec:related-work}
\cutsectiondown

\paragraph{Hierarchical RL.}
A number of hierarchical RL approaches are designed to deal with sequential tasks. Typically these have the form of a meta controller and a set of lower-level controllers for subtasks~\citep{sutton1999between,dietterich2000hierarchical,Parr1997ReinforcementLW,ghavamzadeh2003hierarchical,Konidaris2012TransferIR,Konidaris2007BuildingPO}. However, much of the previous work assumes that the overall task is fixed (e.g., \textit{Taxi} domain ~\citep{dietterich2000hierarchical}). In other words, the optimal sequence of subtasks is fixed during evaluation (e.g., picking up a passenger followed by navigating to a destination in the Taxi domain). This makes it hard to evaluate the agent's ability to compose pre-learned policies to solve previously unseen sequential tasks in a zero-shot fashion unless we re-train the agent on the new tasks in a transfer learning setting~\citep{Singh1991TheEL,singh1992transfer,mcgovern2002autonomous}.
Our work is also closely related to Programmable HAMs (PHAMs)~\citep{Andre2000ProgrammableRL, Andre2002StateAF} in that a PHAM is designed to execute a given program. However, the program explicitly specifies the policy in PHAMs which effectively reduces the state-action search space. In contrast, instructions are a description of the task in our work, which means that the agent should learn to use the instructions to maximize its reward. 

\cutparagraphup
\vspace{-0.1in}
\paragraph{Hierarchical Deep RL.}
Hierarhical RL has been recently combined with deep learning. \citet{kulkarni2016hierarchical} proposed hierarchical Deep Q-Learning and demonstrated improved exploration in a challenging Atari game. \citet{Tessler2017ADH} proposed a similar architecture, but the high-level controller is allowed to choose primitive actions directly. \citet{Bacon2017TheOA} proposed the \textit{option-critic} architecture, which learns options without pseudo reward and demonstrated that it can learn distinct options in Atari games. \citet{Heess2016LearningAT} formulated the actions of the meta controller as continuous variables that are used to modulate the behavior of the low-level controller. \citet{Florensa2017StochasticNN} trained a stochastic neural network with mutual information regularization to discover skills. 
Most of these approaches build an \textit{open-loop} policy at the high-level controller that waits until the previous subtask is finished once it is chosen. This approach is not able to interrupt ongoing subtasks in principle, while our architecture can switch its subtask at any time.

\begin{figure}
    \centering
    \small
	\includegraphics[width=0.70\linewidth]{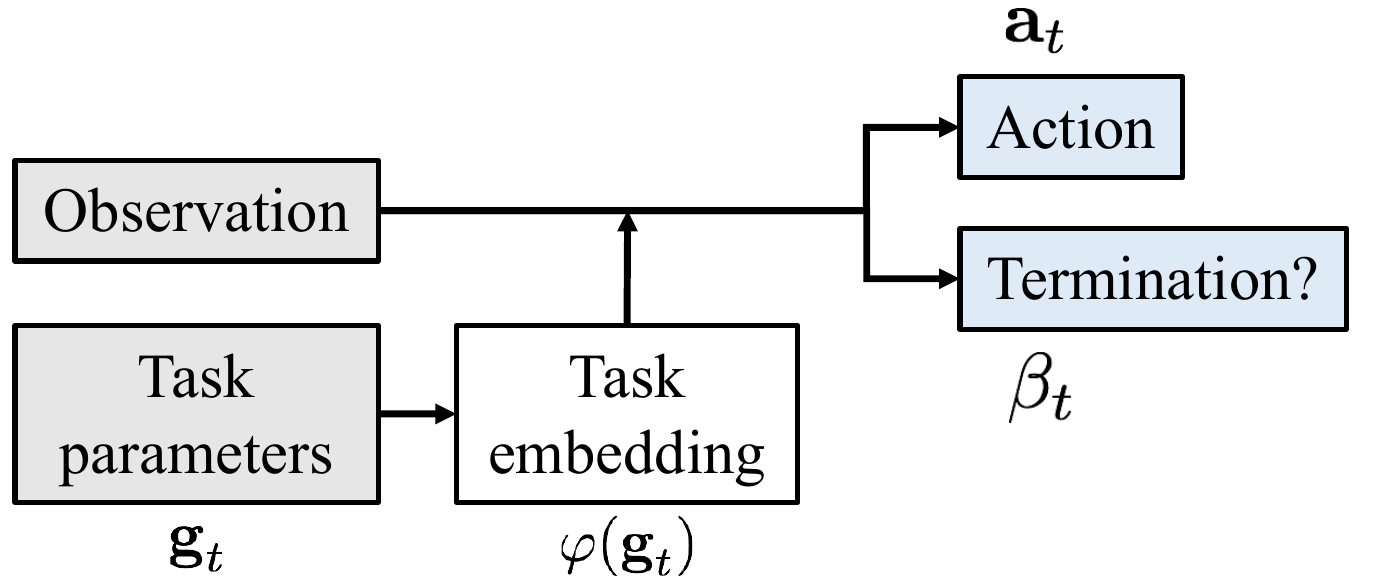} 
	\vspace{-5pt}
	\caption{Architecture of parameterized skill. See text for details.} 
  	\label{fig:arch-multi}
	\vspace{-15pt}
\end{figure}

\cutparagraphup
\paragraph{Zero-Shot Task Generalization.}
There have been a few papers on zero-shot generalization to new tasks. 
For example, \citet{Silva2012LearningPS} introduced parameterized skills that map sets of task descriptions to policies. 
\citet{Isele2016UsingTF} achieved zero-shot task generalization through dictionary learning with sparsity constraints. \citet{schaul2015universal} proposed \textit{universal value function approximators} (UVFAs) that learn  value functions for state and goal pairs. \citet{Devin2016LearningMN} proposed composing sub-networks that are shared across tasks and robots in order to achieve generalization to unseen configurations of them. Unlike the above prior work, we propose a flexible metric learning method which can be applied to various generalization scenarios. \citet{Andreas2016ModularMR} proposed a framework to learn the underlying subtasks from a \textit{policy sketch} which specifies a sequence of subtasks, and the agent can generalize over new sequences of them in principle. In contrast, our work aims to generalize over unseen subtasks as well as unseen sequences of them. In addition, the agent should handle unexpected events in our problem that are not described by the instructions by interrupting subtasks appropriately.

\cutparagraphup
\paragraph{Instruction Execution.}
There has been a line of work for building agents that can execute natural language instructions: \citet{Tellex2011UnderstandingNL,Tellex2014AskingFH} for robotics and \citet{macmahon:aaai06,chen:aaai11,mei2015listen} for a simulated environment. However, these approaches focus on natural language understanding to map instructions to actions or \textit{groundings} in a supervised setting. In contrast, we focus on generalization to sequences of instructions without any supervision for language understanding or for actions. 
Although \citet{Branavan2009ReinforcementLF} also tackle a similar problem, the agent is given a single instruction at a time, while our agent needs to learn how to align instructions and state given a full list of instructions. 

\section{Learning a Parameterized Skill} \label{sec:multi-task} 
\cutsubsectiondown 
In this paper, a parameterized skill is a multi-task policy corresponding to multiple tasks defined by categorical input \textit{task parameters}, e.g., [Pick up, X].
More formally, we define a parameterized skill as a mapping $\mathcal{O} \times \mathcal{G} \rightarrow \mathcal{A} \times \mathcal{B}$, where $\mathcal{O}$ is a set of observations, $\mathcal{G}$ is a set of task parameters, $\mathcal{A}$ is a set of primitive actions, and $\mathcal{B}=\{0,1\}$ indicates whether the task is finished or not. 
A space of tasks is defined using the Cartesian product of task parameters: $\mathcal{G}=\mathcal{G}^{(1)} \times ... \times \mathcal{G}^{(n)}$, 
where $\mathcal{G}^{(i)}$ is a set of the $i$-th parameters (e.g., $\mathcal{G} = $ \{Visit, Pick up\}$\times$\{X, Y, Z\}). 
Given an observation $\textbf{x}_t \in \mathcal{O}$ at time $t$ and task parameters $\textbf{g}=\left[g^{(1)}, ..., g^{(n)}\right] \in \mathcal{G}$, where $g^{(i)}$ is a one-hot vector, the parameterized skill is the following functions: 
\begin{align*}
\mbox{Policy: } & \pi_{\phi}(\textbf{a}_t | \textbf{x}_t, \textbf{g}) \\ \mbox{Termination: } & \beta_{\phi}(b_t | \textbf{x}_t, \textbf{g}),
\end{align*}
where $\pi_{\phi}$ is the policy optimized for the task $\textbf{g}$, and $\beta_{\phi}$ is a termination function~\citep{sutton1999between} which is the probability that the state is terminal at time $t$ for the given task $\textbf{g}$. The parameterized skill is represented by a non-linear function approximator $\phi(\cdot)$, a neural network in this paper. 
The neural network architecture of our parameterized skill is illustrated in Figure~\ref{fig:arch-multi}. The network maps input task parameters into a task embedding space $\varphi(\textbf{g})$, which is combined with the observation followed by the output layers.  More details are described in the \supplementary{}.

\cutsubsectionup
\subsection{Learning to Generalize by Analogy-Making} \label{sec:analogy}
\cutsubsectiondown
Only a subset of tasks ($\mathcal{G'}\subset\mathcal{G}$) are available during training, 
and so in
order to generalize to unseen tasks during evaluation the network needs to learn 
knowledge 
about the relationship between different task parameters when learning the task embedding $\varphi(\textbf{g})$. 

To this end, we propose an analogy-making objective inspired by \citet{reed2015deep}. The main idea is to learn correspondences between tasks. For example, if target objects and `Visit/Pick up' actions are \emph{independent} (i.e., each action can be applied to any target object), we can enforce the analogy [Visit, X] : [Visit, Y] :: [Pick up, X] : [Pick up, Y] for any X and Y in the embedding space, which means that the difference between `Visit' and `Pick up' is consistent regardless of target objects and vice versa. This allows the agent to generalize to unseen combinations of actions and target objects, such as performing [Pick up, Y] after it has learned to perform [Pick up, X] and [Visit, Y].

\newcommand{\diffvec}{\Delta}

More specifically, we define several constraints as follows:
\begin{align*}
\left\Vert \diffvec \left( \textbf{g}_A , \textbf{g}_B \right) - \diffvec \left( \textbf{g}_C,\textbf{g}_D \right) \right\Vert \approx 0 & & \mbox{ if }  \textbf{g}_A : \textbf{g}_B :: \textbf{g}_C : \textbf{g}_D   \\
\left\Vert \diffvec \left( \textbf{g}_A , \textbf{g}_B \right) - \diffvec \left( \textbf{g}_C, \textbf{g}_D \right) \right\Vert \geq \tau_{dis} & &  \mbox{ if }  \textbf{g}_A : \textbf{g}_B \neq \textbf{g}_C : \textbf{g}_D  \\
\left\Vert \diffvec \left( \textbf{g}_A, \textbf{g}_B \right)\right \Vert \geq \tau_{diff} & & \mbox{ if } \textbf{g}_A \neq \textbf{g}_B, 
\end{align*}
where $\textbf{g}_k=\left[ g_k^{(1)}, g_k^{(2)}, ..., g_k^{(n)} \right] \in \mathcal{G}$ are task parameters, $\diffvec \left( \textbf{g}_A , \textbf{g}_B \right)=\varphi(\textbf{g}_A)-\varphi(\textbf{g}_B)$ is the difference vector between two tasks in the embedding space, and $\tau_{dis}$ and $\tau_{diff}$ are constant threshold distances. Intuitively, the first constraint enforces the analogy (i.e., \emph{parallelogram} structure in the embedding space; see~\citet{mikolov2013exploiting,reed2015deep}), while the other constraints prevent trivial solutions. We incorporate these constraints into the following objectives based on contrastive loss~\citep{hadsell2006dimensionality}:
\begin{align*}
& \mathcal{L}_{sim} =  \mathbb{E}_{\textbf{g}_{A...D} \sim \mathcal{G}_{sim}} \left[ \Vert \diffvec \left( \textbf{g}_A,\textbf{g}_B \right) - \diffvec \left( \textbf{g}_C, \textbf{g}_D \right) \Vert^2 \right] \\ 
& \mathcal{L}_{dis} =  \mathbb{E}_{\textbf{g}_{A...D} \sim \mathcal{G}_{dis}} \left[ \left(\tau_{dis} - \Vert \diffvec \left( \textbf{g}_A , \textbf{g}_B \right) - \diffvec \left( \textbf{g}_C , \textbf{g}_D \right) \Vert \right)_{+}^2 \right]  \\
& \mathcal{L}_{diff} =  \mathbb{E}_{\textbf{g}_{A,B} \sim \mathcal{G}_{diff}} \left[ \left(\tau_{diff} - \Vert \diffvec \left( \textbf{g}_A, \textbf{g}_B \right)  \Vert \right)_{+}^2 \right],
\end{align*}
where $(\cdot)_+ = \max(0, \cdot)$ and $\mathcal{G}_{sim},\mathcal{G}_{dis},\mathcal{G}_{diff}$ are sets of task parameters that satisfy corresponding conditions in the above three constraints. The final analogy-making objective is the weighted sum of the above three objectives. 


\cutsubsectionup
\subsection{Training} 
\cutsubsectiondown
The parameterized skill is trained on a set of tasks ($\mathcal{G'} \subset \mathcal{G}$) through the actor-critic method with generalized advantage estimation~\citep{schulman2015high}. 
We also found that pre-training through \textit{policy distillation}~\citep{rusu2015policy,parisotto2015actor} gives slightly better results as discussed in~\citet{Tessler2017ADH}. 
Throughout training, the parameterized skill is also made to predict whether the current state is terminal or not through a binary classification objective, and the analogy-making objective is applied to the task embedding separately. The full details of the learning objectives are described in the \supplementary{}.

\cutsubsectionup
\subsection{Experiments} \label{sec:exp-skill}
\cutsubsectionup
\paragraph{Environment.}
We developed a 3D visual environment using Minecraft based on \citet{oh2016memory} as shown in Figure~\ref{fig:problem}. 
An observation is represented as a $64 \times 64$ pixel RGB image. There are 7 different types of objects: \textit{Pig}, \textit{Sheep}, \textit{Greenbot}, \textit{Horse}, \textit{Cat}, \textit{Box}, and \textit{Ice}. The topology of the world and the objects are randomly generated for every episode. 
The agent has 9 actions: \textit{Look} (Left/Right/Up/Down), \textit{Move} (Forward/Backward), \textit{Pick up}, \textit{Transform}, and \textit{No operation}. 
\textit{Pick up} removes the object in front of the agent, and \textit{Transform} changes the object in front of the agent to ice (a special object). 

\cutparagraphup
\paragraph{Implementation Details.}
\cutparagraphdown
The network architecture of the parameterized skill consists of 4 convolution layers and one LSTM~\citep{hochreiter1997long} layer. 
We conducted curriculum training by changing the size of the world, the density of object and walls according to the agent's success rate.
We implemented actor-critic method with 16 CPU threads based on~\citet{sukhbaatar2015mazebase}. The parameters are updated after $8$ episodes for each thread. The details of architectures and hyperparameters are described in the \supplementary{}. 

\begin{table}
\centering
\small
\setlength{\tabcolsep}{4pt}
\begin{tabular}{c|c|c|c}
  \hline
Scenario & Analogy & Train & Unseen \\ \hline 
\multirow{2}{*}{Independent} & $\times$ & \textbf{0.3} $(99.8\%)$ & -3.7 $(34.8\%)$ \\
& \checkmark & \textbf{0.3} $(99.8\%)$ & \textbf{0.3} $(99.5\% )$ \\ \hline
\multirow{2}{*}{Object-dependent} & $\times$ & \textbf{0.3} $(99.7\%)$ & -5.0 $(2.2\%)$ \\
& \checkmark & \textbf{0.3} $(99.8\%)$ & \textbf{0.3} $(99.7\%)$ \\ \hline
\multirow{2}{*}{Inter/Extrapolation} & $\times$ & \textbf{-0.7} $(97.5\%)$ & -2.2 $(24.9\%)$ \\
& \checkmark & \textbf{-0.7} $(97.5\%)$ & \textbf{-1.7} $(94.5\% )$ \\ \hline
\end{tabular}
  \vspace{-5pt}
  \caption{Performance on parameterized tasks. Each entry shows `Average reward (Success rate)'. We assume an episode is successful only if the agent successfully finishes the task and its termination predictions are correct throughout the whole episode. }
  \label{tab:subtask}
  \vspace{-15pt}
\end{table}

\cutparagraphup
\paragraph{Results.} 
\cutparagraphdown
To see how useful analogy-making is for generalization to unseen parameterized tasks, we trained and evaluated the parameterized skill on three different sets of parameterized tasks defined below\footnote{The sets of subtasks used for training and evaluation are described in the \supplementary{}.}.
\vspace{-8pt}
\begin{itemize}[leftmargin=*]
\setlength\itemsep{0em}
\item \textbf{Independent}: The task space is defined as $\mathcal{G}=\mathcal{T} \times \mathcal{X}$, where $\mathcal{T}=\{\mbox{Visit}, \mbox{Pick up}, \mbox{\Hit{}}\}$ and $\mathcal{X}$ is the set of object types. The agent should move on top of the target object given `Visit' task and perform the corresponding actions in front of the target given `Pick up' and `Transform' tasks. Only a subset of tasks are encountered during training, so the agent should generalize over unseen configurations of task parameters.
\item \textbf{Object-dependent}: The task space is defined as $\mathcal{G}=\mathcal{T}' \times \mathcal{X}$, where $\mathcal{T}' = \mathcal{T} \cup \left\{\mbox{Interact with}\right\}$. We divided objects into two groups, each of which should be either picked up or transformed given `Interact with' task. Only a subset of target object types are encountered during training, so there is no chance for the agent to generalize without knowledge of the group of each object. We applied analogy-making so that analogies can be made only within the same group. 
This allows the agent to perform object-dependent actions even for unseen objects. 
\item \textbf{Interpolation/Extrapolation}: The task space is defined as $\mathcal{G}=\mathcal{T} \times \mathcal{X} \times \mathcal{C}$, where $\mathcal{C}=\{1,2,...,7\}$. The agent should perform a task for a given number of times ($c\in\mathcal{C}$). Only $\{1,3,5\}\subset \mathcal{C}$ is given during training, and the agent should generalize over unseen numbers $\{2,4,6,7\}$. Note that the optimal policy for a task can be derived from $\mathcal{T} \times \mathcal{X}$, but predicting termination requires generalization to unseen numbers. We applied analogy-making based on arithmetic (e.g., [Pick up, X, 2] : [Pick up, X, 5] :: [Transform, Y, 3] : [Transform, Y, 6]).
\end{itemize}

As summarized in Table~\ref{tab:subtask}, the parameterized skill with our analogy-making objective can successfully generalize to unseen tasks in all generalization scenarios. This suggests that when learning a representation of task parameters, it is possible to inject prior knowledge in the form of the analogy-making objective so that the agent can learn to generalize over unseen tasks in various ways depending on semantics or context without needing to experience them. 

\begin{figure}
		\centering
\includegraphics[width=0.65\linewidth]{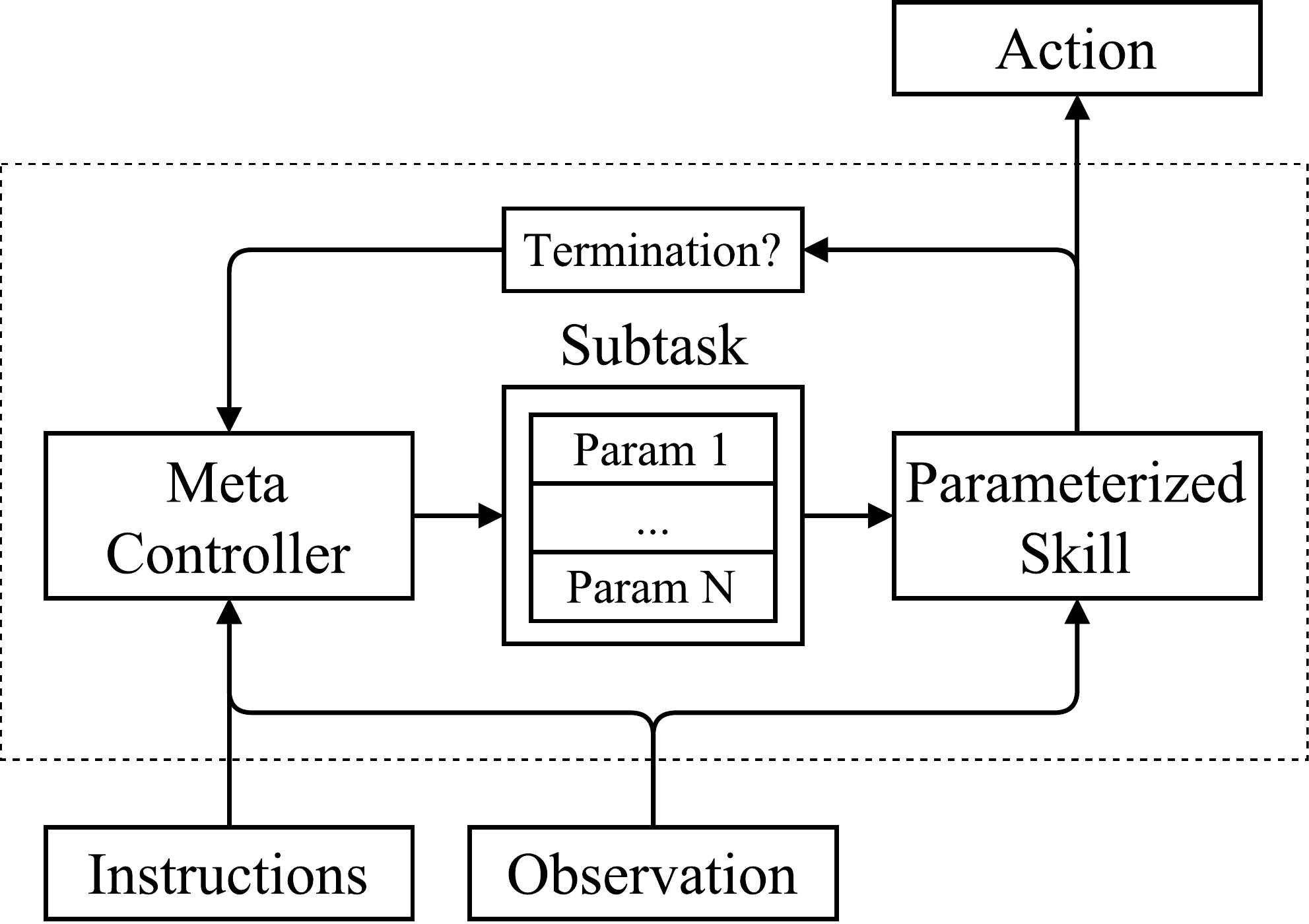} 
        \vspace{-5pt}
  	    \caption{Overview of our hierarchical architecture.} 
		\vspace{-15pt}
  	    \label{fig:overview}
\end{figure}

\cutsectionup
\section{Learning to Execute Instructions using Parameterized Skill} \label{sec:meta}
\cutsectiondown
We now consider the instruction execution problem where the agent is given a sequence of simple natural language instructions, as illustrated in Figure~\ref{fig:problem}.
We assume an already trained parameterized skill, as described in Section~\ref{sec:multi-task}. 
Thus, the main remaining problem is how to use the parameterized skill to execute instructions. Although the requirement that instructions be executed sequentially makes the problem easier (than, e.g., conditional-instructions), the agent still needs to make complex decisions because it should deviate from instructions to deal with unexpected events (e.g., low battery) and remember what it has done to deal with loop instructions, as discussed in Section~\ref{sec:Introduction}.

To address the above challenges, our hierarchical RL architecture (see Figure~\ref{fig:overview}) consists of two modules: meta controller and parameterized skill. Specifically, a meta controller reads the instructions and passes subtask parameters to a parameterized skill which executes the given subtask and provides its termination signal back to the meta controller. 
Section~\ref{sec:meta-archtecture} describes the overall architecture of the meta controller for dealing with instructions. Section~\ref{sec:temporal-abstraction} describes a novel neural architecture that learns when to update the subtask in order to better deal with delayed reward signal as well as unexpected events. 

\cutsubsectionup
\subsection{Meta Controller Architecture} \label{sec:meta-archtecture}
\begin{figure}
    \centering
    \small
	\includegraphics[width=0.9\linewidth]{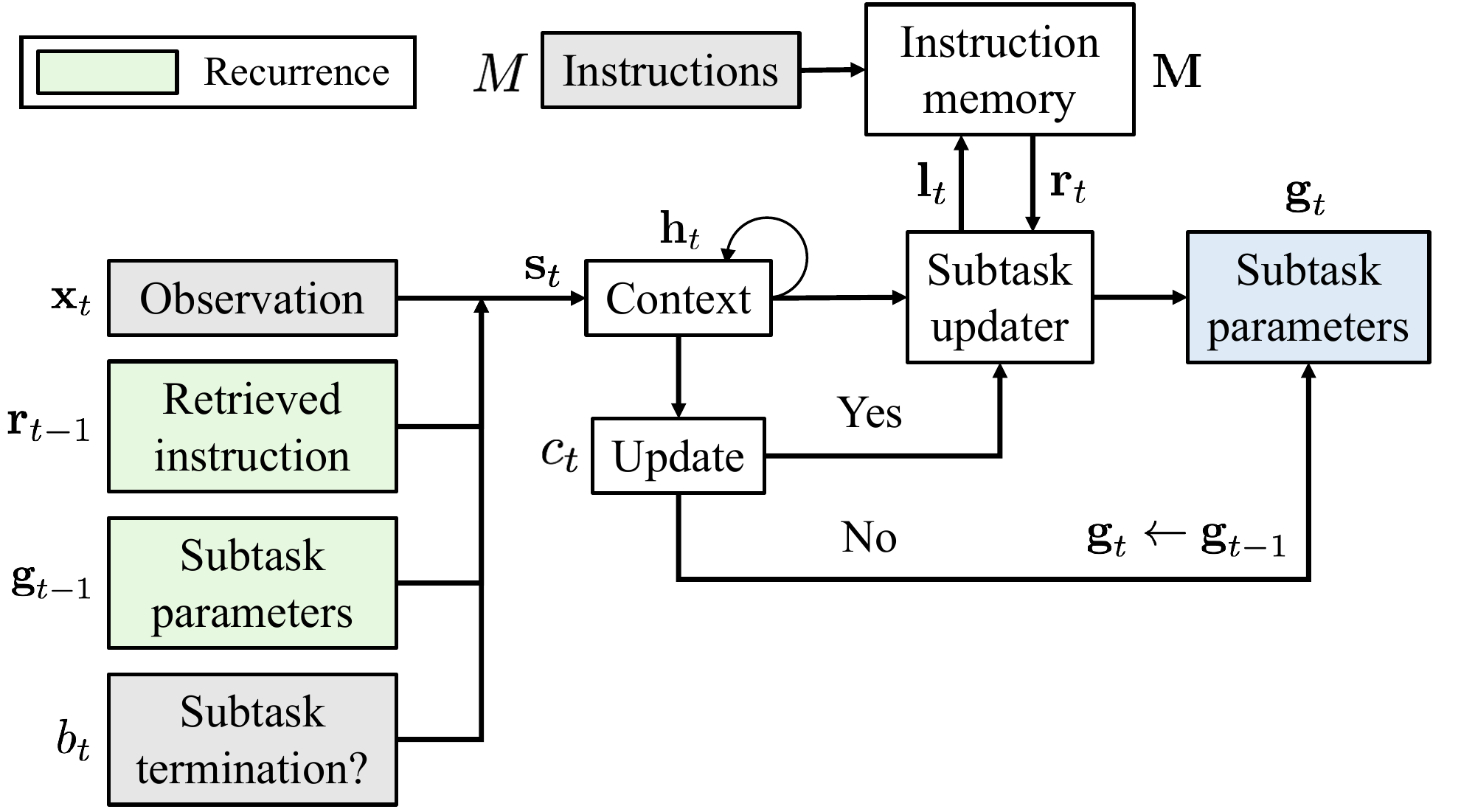} 
	\vspace{-5pt}
	\caption{Neural network architecture of meta controller.} 
	\label{fig:arch-meta}
	\vspace{-15pt}
\end{figure}

\cutsubsectiondown
As illustrated in Figure~\ref{fig:arch-meta}, the meta controller is a mapping $\mathcal{O} \times \mathcal{M} \times \mathcal{G} \times \mathcal{B} \rightarrow \mathcal{G}$, where $\mathcal{M}$ is a list of instructions. Intuitively, the meta controller decides subtask parameters $\textbf{g}_t \in \mathcal{G}$ conditioned on the observation $\textbf{x}_t \in \mathcal{O}$, the list of instructions $M\in \mathcal{M}$, the previously selected subtask $\textbf{g}_{t-1}$, and its termination signal ($b \sim \beta_{\phi}$). 

In contrast to recent hierarchical deep RL approaches where the meta controller can update its subtask (or option) only when the previous one terminates or only after a fixed number of steps, our meta controller can update the subtask at any time and takes the termination signal as additional input. This gives more flexibility to the meta controller and enables interrupting ongoing tasks before termination.

In order to keep track of the agent's progress on instruction execution, the meta controller maintains its internal state by computing a \textit{context} vector (Section~\ref{sec:context}) and determines which subtask to execute by focusing on one instruction at a time from the list of instructions (Section~\ref{sec:subtask-selector}). 

\cutsubsectionup
\subsubsection{Context} \label{sec:context}
\cutsubsectiondown
Given the sentence embedding $\textbf{r}_{t-1}$ retrieved at the previous time-step from the instructions (described in Section~\ref{sec:subtask-selector}), the previously selected subtask $\textbf{g}_{t-1}$, and the subtask termination $b_t \sim \beta_{\phi}\left(b_t\vert\textbf{s}_t,\textbf{g}_{t-1}\right)$, the meta controller computes the context vector ($\textbf{h}_t$) as follows:
\begin{align*}
 \textbf{h}_t & = \mbox{LSTM}\left(\textbf{s}_t, \textbf{h}_{t-1} \right) \\ \textbf{s}_{t} & = f\left(\textbf{x}_{t},\textbf{r}_{t-1},\textbf{g}_{t-1},b_t\right),
\end{align*} 
where $f$ is a neural network. 
Intuitively, $\textbf{g}_{t-1}$ and $b_t$ provide information about which subtask was being solved by the parameterized skill and whether it has finished or not. Thus, $\textbf{s}_t$ is a summary of the current observation and the ongoing subtask. $\textbf{h}_t$ takes the history of $\textbf{s}_t$ into account through the LSTM, which is used by the subtask updater. 

\cutsubsectionup
\subsubsection{Subtask Updater} \label{sec:subtask-selector}
\cutsubsectiondown
The \textit{subtask updater} constructs a memory structure from the list of instructions, retrieves an instruction by maintaining a pointer into the memory, and computes the subtask parameters.

\cutparagraphup
\paragraph{Instruction Memory.} Given instructions as a list of sentences $M=\left(\mathbf{m}_1,\mathbf{m}_2,...,\mathbf{m}_K \right)$, where each sentence consists of a list of words, $\mathbf{m}_i=\left(w_1,...,w_{|\mathbf{m}_i|}\right)$, the subtask updater constructs memory blocks $\textbf{M} \in \mathbb{R}^{E \times K}$ (i.e., each column is an $E$-dimensional embedding of a sentence).  
The subtask updater maintains an \textit{instruction pointer} ($\textbf{p}_t \in \mathbb{R}^{K}$) which is non-negative and sums up to 1 indicating which instruction the meta controller is executing.
Memory construction and retrieval can be written as:
\begin{align}
\mbox{Memory: }\textbf{M} & = \left[\varphi^{w}\left(\mathbf{m}_1\right),\varphi^{w}\left(\mathbf{m}_2\right),...,\varphi^{w}\left(\mathbf{m}_K\right) \right] \\
\mbox{Retrieval: } \textbf{r}_t & = \textbf{M}\textbf{p}_t, \label{eq:retrieval}
\end{align}
where $\varphi^{w}\left(\mathbf{m}_i\right) \in \mathbb{R}^{E}$ is the embedding of the $i$-th sentence (e.g., Bag-of-words), and $\textbf{r}_t \in \mathbb{R}^{E}$ is the retrieved sentence embedding which is used for computing the subtask parameters. Intuitively, if $\textbf{p}_t$ is a one-hot vector, $\textbf{r}_t$ indicates a single instruction from the whole list of instructions. The meta controller should learn to manage  $\textbf{p}_t$ so that it can focus on the correct instruction at each time-step.

Since instructions should be executed sequentially, we use a location-based memory addressing mechanism~\citep{zaremba2015reinforcement,graves2014neural} to manage the instruction pointer. Specifically, the subtask updater shifts the instruction pointer by $[-1,1]$ as follows:
\begin{align}
\textbf{p}_t = \textbf{l}_t \ast \textbf{p}_{t-1} \mbox{ where } \textbf{l}_t = \mbox{Softmax}\left( \varphi^{shift}(\textbf{h}_t) \right), \label{eq:shift}
\end{align}
where $\ast$ is a convolution operator, $\varphi^{shift}$ is a neural network, and $\textbf{l}_t \in \mathbb{R}^3$ is a soft-attention vector over the three shift operations $\{-1, 0, +1\}$. The optimal policy should keep the instruction pointer unchanged while executing an instruction and increase the pointer by +1 precisely when the current instruction is finished.

\cutparagraphup
\paragraph{Subtask Parameters.} 
The subtask updater takes the context ($\textbf{h}_t$), updates the instruction pointer ($\textbf{p}_t$), retrieves an instruction ($\textbf{r}_t$), and computes subtask parameters as:
\begin{align}
\pi_{\theta}\left(\textbf{g}_t \vert \textbf{h}_t, \textbf{r}_t \right) = \prod_i \pi_{\theta}\left(g^{(i)}_t \vert \textbf{h}_t, \textbf{r}_t \right), \label{eq:subtask}
\end{align}
where $\pi_{\theta}\left(g^{(i)}_t \vert \textbf{h}_t, \textbf{r}_t \right) \propto \exp \left( \varphi^{goal}_i\left(\textbf{h}_t, \textbf{r}_t \right) \right)$, and $\varphi^{goal}_i$ is a neural network for the $i$-th subtask parameter. 


\begin{figure}
    \centering
    \small
	\includegraphics[width=0.95\linewidth]{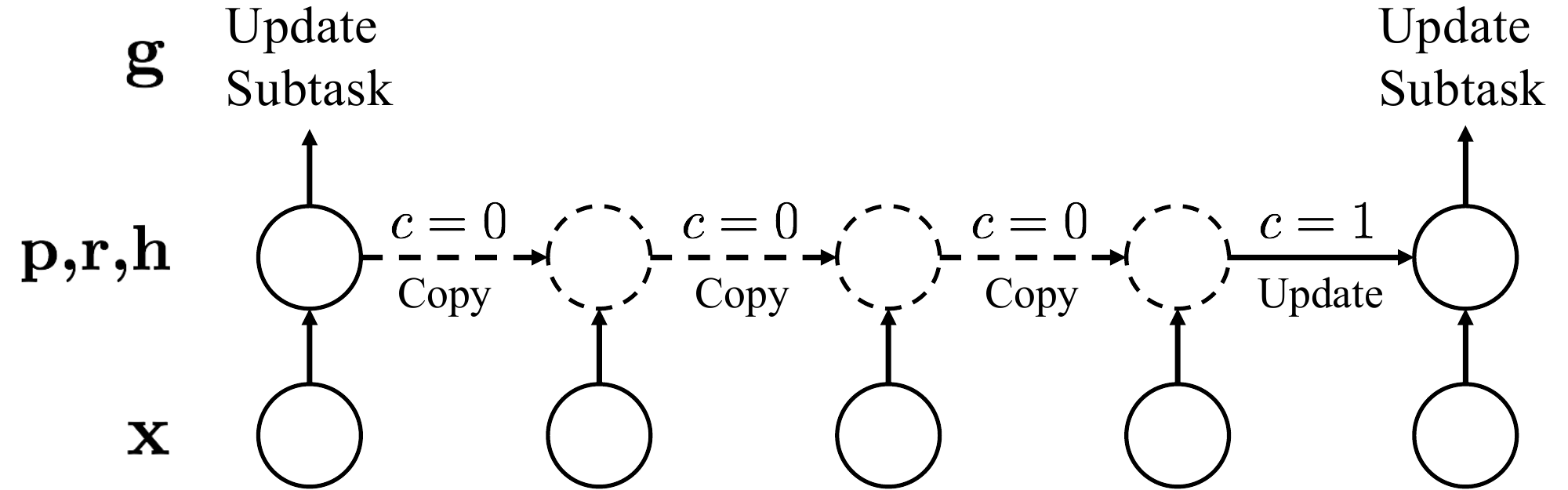} 
	\vspace{-5pt}
	\caption{Unrolled illustration of the meta controller with a learned time-scale. The internal states ($\textbf{p},\textbf{r},\textbf{h}$) and the subtask ($\textbf{g}$) are updated only when $c=1$. If $c=0$, the meta controller continues the previous subtask without updating its internal states. } 
  	\label{fig:ta}
	\vspace{-10pt}
\end{figure}
\cutsubsectionup
\subsection{Learning to Operate at a Large Time-Scale} \label{sec:temporal-abstraction}
\cutsubsectiondown
Although the meta controller can learn an optimal policy by updating the subtask at each time-step in principle, making a decision at every time-step can be inefficient because subtasks do not change frequently. Instead, having temporally-extended actions can be useful for dealing with delayed reward by operating at a larger time-scale~\citep{sutton1999between}. 
While it is reasonable to use the subtask termination signal to define the temporal scale of the meta controller as in many recent hierarchical deep RL approaches (see Section~\ref{sec:related-work}), this approach would result in a mostly open-loop meta-controller policy that is not able to interrupt ongoing subtasks before termination, which is necessary to deal with unexpected events not specified in the instructions. 

To address this dilemma, we propose to learn the time-scale of the meta controller by introducing an internal binary decision which indicates whether to \textit{invoke} the subtask updater to update the subtask or not, as illustrated in Figure~\ref{fig:ta}.
This decision is defined as: 
$c_t \sim \sigma \left( \varphi^{update}\left(\textbf{s}_t,\textbf{h}_{t-1} \right) \right)$ where $\sigma$ is a sigmoid function.
If $c_t=0$, the meta controller continues the current subtask without updating the subtask updater. Otherwise, if $c_t=1$, the subtask updater updates its internal states (e.g., instruction pointer) and the subtask parameters. This allows the subtask updater to operate at a large time-scale because one decision made by the subtask updater results in multiple actions depending on $c$ values. The overall meta controller architecture with this update scheme is illustrated in Figure~\ref{fig:arch-meta}. 

\cutsubsectionup
\paragraph{Soft-Update.}
\cutsubsectiondown
To ease optimization of the non-differentiable variable ($c_t$), we propose a \textit{soft-update} rule by using $c_t=\sigma\left( \varphi^{update}\left(\textbf{s}_t,\textbf{h}_{t-1} \right) \right)$ instead of sampling it. The key idea is to take the weighted sum of both `update' and `copy' scenarios using $c_t$ as the weight. This method is described in Algorithm~\ref{alg2}. We found that training the meta controller using soft-update followed by fine-tuning by sampling $c_t$ is crucial for training the meta controller. Note that the  soft-update rule reduces to the original formulation if we sample $c_t$ and $\textbf{l}_t$ from the Bernoulli and multinomial distributions, which justifies our initialization trick.

\begin{center}
\begin{minipage}{1\linewidth}
\vspace{-15pt}
\begin{algorithm}[H]
\small
\caption{Subtask update (Soft)}\label{alg2}
\begin{algorithmic}
\STATE \textbf{Input:} $\textbf{s}_{t},\textbf{h}_{t-1},\textbf{p}_{t-1},\textbf{r}_{t-1},\textbf{g}_{t-1}$
\STATE \textbf{Output:} $\textbf{h}_{t},\textbf{p}_{t},\textbf{r}_t,\textbf{g}_t$
\STATE $c_t \gets \sigma \left( \varphi^{update}\left(\textbf{s}_t,\textbf{h}_{t-1} \right) \right)$ \hfill \# Decide update weight
\STATE $\tilde{\textbf{h}}_t \gets \mbox{LSTM} \left(\textbf{s}_{t}, \textbf{h}_{t-1} \right) $ \hfill \# Update the context
\STATE $\textbf{l}_t \gets \mbox{Softmax} \left( \varphi^{shift}\left(\tilde{\textbf{h}}_t \right) \right) $ \hfill \# Decide shift operation
\STATE $\tilde{\textbf{p}}_t \gets \textbf{l}_t \ast  \textbf{p}_{t-1} $  \hfill \# Shift the instruction pointer
\STATE $\tilde{\textbf{r}}_t \gets \textbf{M}\tilde{\textbf{p}}_t $ \hfill \# Retrieve instruction
\STATE \# Merge two scenarios (update/copy) using $c_t$ as weight
\STATE $[\textbf{p}_t, \textbf{r}_t, \textbf{h}_t] \gets c_t [\tilde{\textbf{p}}_t, \tilde{\textbf{r}}_t, \tilde{\textbf{h}}_t] + \left(1 - c_t\right)[\textbf{p}_{t-1}, \textbf{r}_{t-1}, \textbf{h}_{t-1}]$ 
\STATE $g^{(i)}_t \sim c_t \pi_{\theta}\left(g^{(i)}_t|\tilde{\textbf{h}}_t,\tilde{\textbf{r}}_t\right) + \left(1-c_t\right)g^{(i)}_{t-1} \forall i $
\end{algorithmic}
\end{algorithm}
\vspace{-15pt}
\end{minipage}
\end{center}

\cutsubsectionup
\paragraph{Integrating with Hierarchical RNN.}
\cutsubsectiondown
The idea of learning the time-scale of a recurrent neural network is closely related to hierarchical RNN approaches~\citep{Koutnk2014ACR,Chung2016HierarchicalMR} where different groups of recurrent hidden units operate at different time-scales to capture both long-term and short-term temporal information. Our idea can be naturally integrated with hierarchical RNNs by applying the update decision ($c$ value) only for a subset of recurrent units instead of all the units. Specifically, we divide the context vector into two groups: $\textbf{h}_t=\left[\textbf{h}^{(l)}_t,\textbf{h}^{(h)}_t\right]$. The low-level units ($\textbf{h}^{(l)}_t$) are updated at every time-step, while the high-level units ($\textbf{h}^{(h)}_t$) are updated depending on the value of $c$. This simple modification leads to a form of hierarchical RNN where the low-level units focus on short-term temporal information while the high-level units capture long-term dependencies.

\cutsubsectionup
\vspace{-0.05in}
\subsection{Training} 
\cutsubsectiondown
The meta controller is trained on a training set of lists of instructions. Given a pre-trained and fixed parameterized skill, the actor-critic method is used to update the parameters of the meta controller. Since the meta controller also learns a subtask embedding $\varphi(\textbf{g}_{t-1})$ and has to deal with unseen subtasks during evaluation, analogy-making objective is also applied.
The details of the objective function are provided in the \supplementary{}.  

\cutsectionup
\vspace{-0.05in}
\subsection{Experiments} \label{sec:exp-meta}
\cutsectiondown
The experiments are designed to explore the following questions: (1) Will the proposed hierarchical architecture outperform a non-hierarchical baseline? (2) How beneficial is the meta controller's ability to learn when to update the subtask? We are also interested in understanding the qualitative properties of our agent's behavior.\footnote{For further analysis, we also conducted comprehensive experiments on a 2D grid-world domain. However, due to space limits, those results are provided in the \supplementary{}.} 

\cutparagraphup
\paragraph{Environment.}
\cutparagraphdown
We used the same Minecraft domain used in Section~\ref{sec:exp-skill}. The agent receives a time penalty ($-0.1$) for each step and receives $+1$ reward when it finishes the entire list of instructions in the correct order. Throughout an episode, a box (including treasures) randomly appears with probability of $0.03$ and \hitting{} a box gives $+0.9$ reward. 

The subtask space is defined as $\mathcal{G}=\mathcal{T} \times \mathcal{X}$, and the semantics of each subtask are the same as the `Independent' case in Section~\ref{sec:exp-skill}. 
We used the best-performing parameterized skill throughout this experiment.

There are 7 types of instructions: \{Visit X, Pick up X, \Hit{} X, Pick up 2 X, \Hit{} 2 X, Pick up 3 X, \Hit{} 3 X\} where `X' is the target object type. Note that the parameterized skill used in this experiment was not trained on loop instructions (e.g., Pick up 3 X), so the last four instructions require the meta controller to learn to repeat the corresponding subtask for the given number of times. To see how the agent generalizes to previously unseen instructions, only a subset of instructions and subtasks was presented during training.

\cutparagraphup
\paragraph{Implementation Details.}
\cutparagraphdown
The meta controller consists of 3 convolution layers and one LSTM layer. 
We also conducted curriculum training by changing the size of the world, the density of object and walls, and the number of instructions according to the agent's success rate. 
We used the actor-critic implementation described in Section~\ref{sec:exp-skill}.
\begin{table}
\centering
\small
\setlength{\tabcolsep}{2.4pt}
\begin{tabular}{r|c|c|c}
\hline
& Train & Test (Seen) & Test (Unseen)  \\ 
\hline
Length of instructions & 4 & 20 & 20 \\ \hline
 Flat & -7.1 $(1\%)$ & -63.6 $(0\%)$ & -62.0 $(0\%)$ \\ 
 Hierarchical-Long & -5.8 $(31\%)$ & -59.2 $(0\%)$ & -59.2 $(0\%)$  \\  
 Hierarchical-Short & -3.3 $(83\%)$ & -53.4 $(23\%)$ & -53.6 $(18\%)$  \\ \hline
 \textbf{Hierarchical-Dynamic} & \textbf{-3.1} $(95\%)$ & \textbf{-30.3} $(75\%)$ & \textbf{-38.0} $(56\%)$ \\ \hline 
\end{tabular}
\vspace{-5pt}
\caption{Performance on instruction execution. Each entry shows average reward and success rate. `Hierarchical-Dynamic' is our approach that learns when to update the subtask. An episode is successful only when the agent solves all instructions correctly. }
\label{tab:performance}
\vspace{-15pt}
\end{table}

\newcommand{\figthreesubplotlen}{0.18\textwidth}
\newcommand{\legendlen}{0.2\textwidth}
\begin{figure*}
    \centering
   	\hspace{-5pt}
    \begin{subfigure}{\figthreesubplotlen}
    	\centering
	    \includegraphics[width=\textwidth]{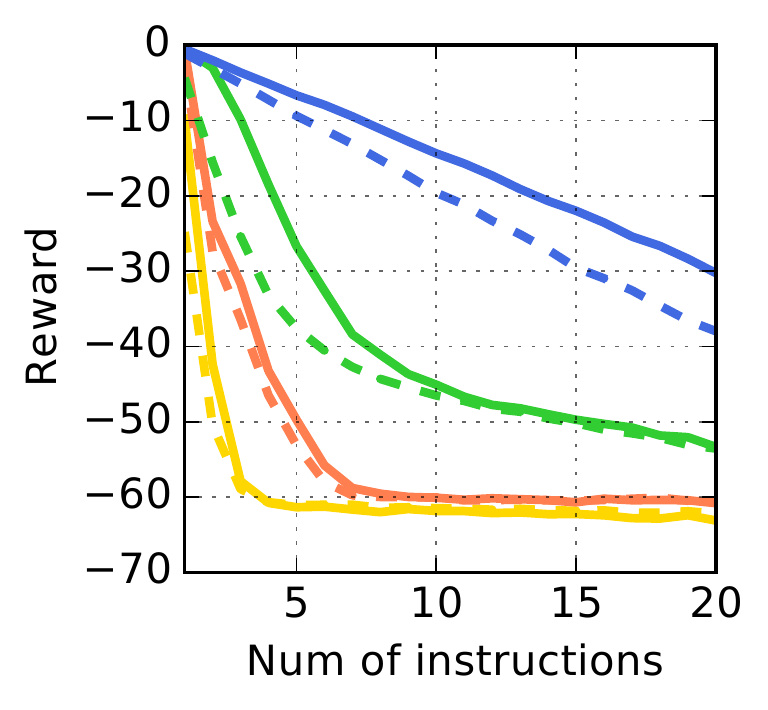} 
   	\end{subfigure}
   	\hspace{-7pt}
	\begin{subfigure}{\figthreesubplotlen}
		\centering
	    \includegraphics[width=\textwidth]{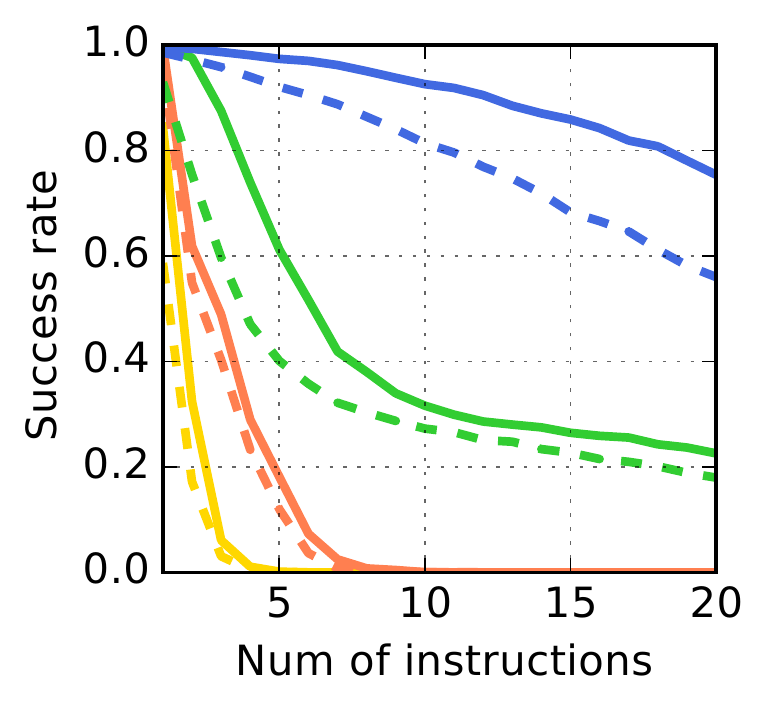} 
	\end{subfigure}
   	\hspace{-7pt}
	\begin{subfigure}{\figthreesubplotlen}
		\centering
	    \includegraphics[width=\textwidth]{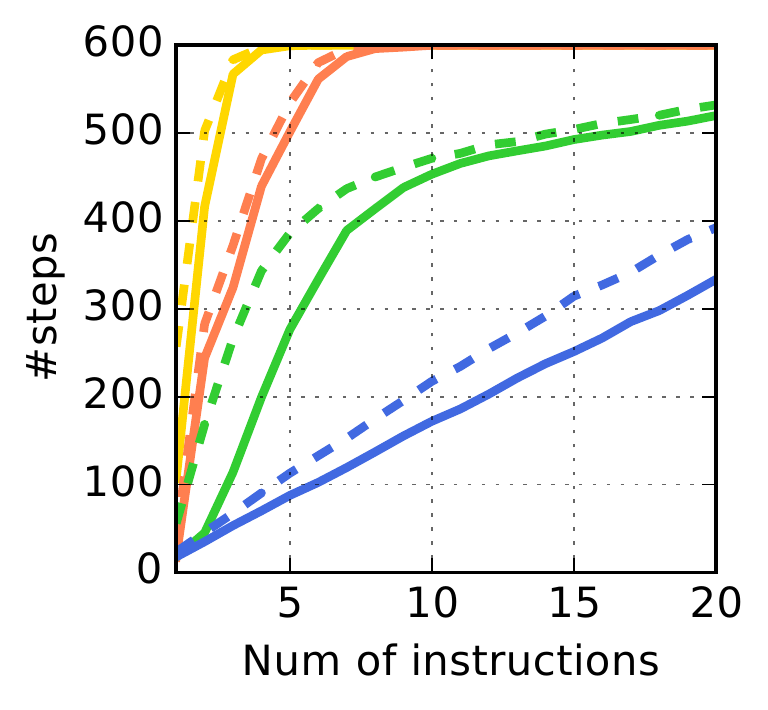} 
	\end{subfigure}
   	\hspace{-7pt}
	\begin{subfigure}{\figthreesubplotlen}
		\centering
	    \includegraphics[width=\textwidth]{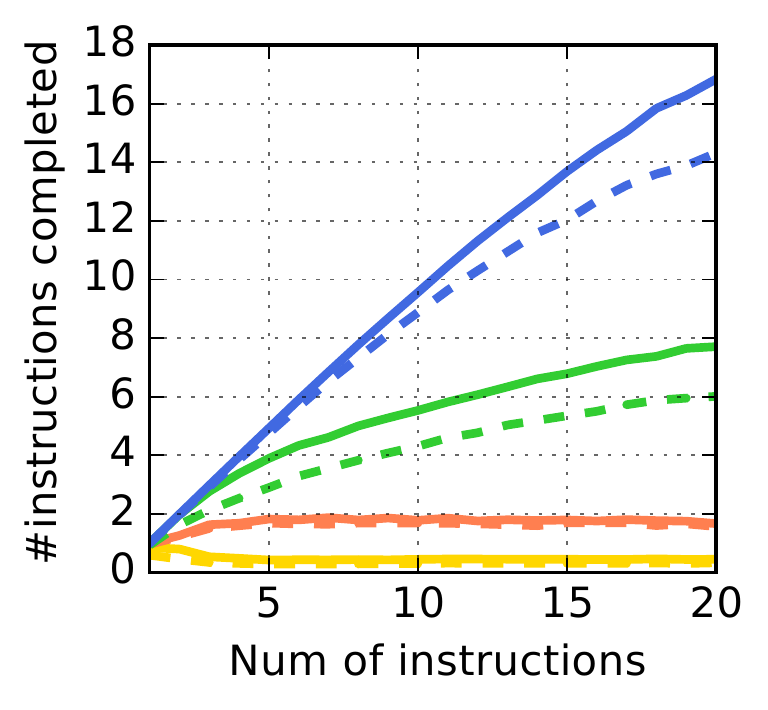} 
	\end{subfigure}
   	\hspace{-5pt}
	\begin{subfigure}{\legendlen}
		\centering
	    \raisebox{3mm}{\includegraphics[width=\textwidth]{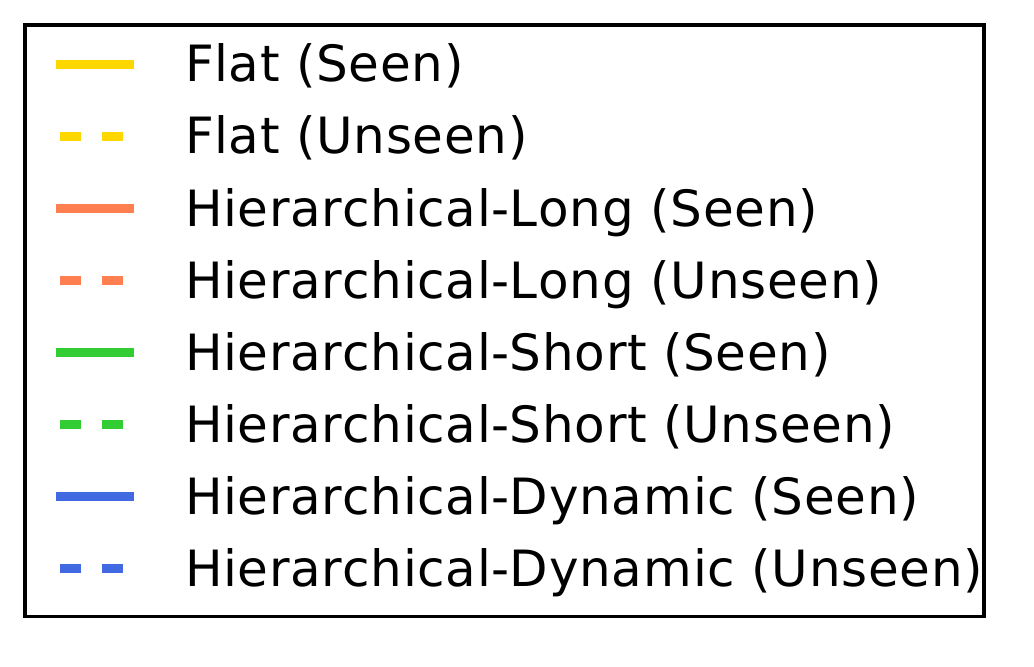}}
	\end{subfigure}
	\vspace{-5pt}
	\caption{Performance per number of instructions. From left to right, the plots show reward, success rate, the number of steps, and the average number of instructions completed respectively.} 
	\vspace{-10pt}
	\label{fig:curve}
\end{figure*}

\begin{figure*}
    \centering
    \includegraphics[width=0.90\textwidth]{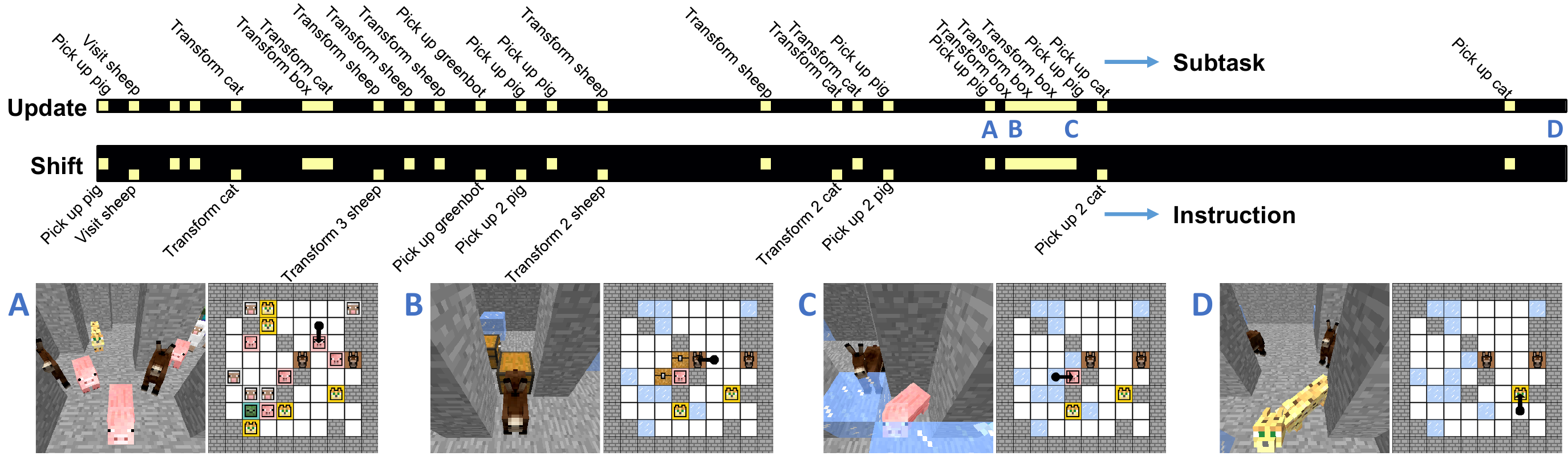} 
    \vspace{-5pt}
	\caption{Analysis of the learned policy. `Update' shows our agent's internal update decision. `Shift' shows our agent's instruction-shift decision (-1, 0, and +1 from top to bottom). The bottom text shows the instruction indicated by the instruction pointer, while the top text shows the subtask chosen by the meta controller. (A) the agent picks up the pig to finish the instruction and moves to the next instruction. (B) When the agent observes a box that randomly appeared while executing `Pick up 2 pig' instruction, it immediately changes its subtask to [\Hit{}, Box]. (C) After dealing with the event (transforming a box), the agent resumes executing the instruction (`Pick up 2 pig'). (D) The agent finishes the final instruction. }
	\label{fig:analysis}
	\vspace{-10pt}
\end{figure*}

\cutparagraphup
\paragraph{Baselines.}
To understand the advantage of using the hierarchical structure and the benefit of our meta controller's ability to learn when to update the subtask, we trained three baselines as follows.
\vspace{-8pt}
\begin{itemize}[leftmargin=*]
\setlength\itemsep{0em}
\item \textbf{Flat}: identical to our meta controller except that it directly chooses primitive actions without using the parameterized skill. It is also pre-trained on the training set of subtasks. 
\item \textbf{Hierarchical-Long}: identical to our architecture except that the meta controller can update the subtask only when the current subtask is finished. This approach is similar to recent hierarchical deep RL methods~\citep{kulkarni2016hierarchical,Tessler2017ADH}.
\item \textbf{Hierarchical-Short}: identical to our architecture except that the meta controller updates the subtask at every time-step.
\end{itemize}
\vspace{-10pt}


\cutparagraphup
\paragraph{Overall Performance.}
The results on the instruction execution are summarized in Table~\ref{tab:performance} and Figure~\ref{fig:curve}. It shows that our architecture (`\textbf{Hierarchical-Dynamic}') can handle a relatively long list of seen and unseen instructions of length 20 with reasonably high success rates, even though it is trained on short instructions of length 4.
Although the performance degrades as the number of instructions increases, our architecture finishes 18 out of 20 seen instructions and 14 out of 20 unseen instructions on average.
These results show that our agent is able to generalize to longer compositions of seen/unseen instructions by just learning to solve short sequences of a subset of instructions. 

\cutparagraphup
\paragraph{Flat vs. Hierarchy.} 
Table~\ref{tab:performance} shows that the flat baseline completely fails even on training instructions. The flat controller tends to struggle with loop instructions (e.g., Pick up 3 pig) so that it learned a sub-optimal policy which moves to the next instruction with a small probability at each step regardless of its progress. This implies that it is hard for the flat controller to detect precisely when a subtask is finished, whereas hierarchical architectures can easily detect when a subtask is done, because the parameterized skill provides a termination signal to the meta controller.

\cutparagraphup
\paragraph{Effect of Learned Time-Scale.}
As shown in Table~\ref{tab:performance} and Figure~\ref{fig:curve}, `Hierarchical-Long' baseline performs significantly worse than our architecture. We found that whenever a subtask is finished, this baseline puts a high probability to switch to [Transform, Box] regardless of the existence of box because transforming a box gives a bonus reward if a box exists by chance. However, this leads to wasting too much time finding a box until it appears and results in a poor success rate due to the time limit. This result implies that an open-loop policy that has to wait until a subtask finishes can be confused by such an uncertain event because it cannot interrupt ongoing subtasks before termination. 

On the other hand, we observed that `Hierarchical-Short' often fails on loop instructions by moving on to the next instruction before it finishes such instructions. This baseline should repeat the same subtask while not changing the instruction pointer for a long time and the reward is even more delayed given loop instructions. In contrast, the subtask updater in our architecture makes fewer decisions by operating at a large time-scale so that it can get more direct feedback from the long-term future. We conjecture that this is why our architecture performs better than this baseline. This result shows that learning when to update the subtask using the neural network is beneficial for dealing with delayed reward without compromising the ability to interrupt.
	    
\cutparagraphup
\vspace{-0.1in}
\paragraph{Analysis of The Learned Policy.}
We visualized our agent's behavior given a long list of instructions in Figure~\ref{fig:analysis}. Interestingly, when the agent sees a box, the meta controller immediately changes its subtask to [\Hit{}, Box] to get a positive reward even though its instruction pointer is indicating `Pick up 2 pig' and resumes executing the instruction after dealing with the box. Throughout this event and the loop instruction, the meta controller keeps the instruction pointer unchanged as illustrated in (B-C) in Figure~\ref{fig:analysis}. 
In addition, the agent learned to update the instruction pointer and the subtask almost only when it is needed, which provides the subtask updater with temporally-extended actions. This is not only computationally efficient but also useful for learning a better policy.

\cutsectionup
\section{Conclusion}
\cutsectiondown
In this paper, we explored a type of zero-shot task generalization in RL with a new problem where the agent is required to execute and generalize over sequences of instructions. We proposed an analogy-making objective which enables generalization over unseen parameterized tasks in various scenarios. We also proposed a novel way to learn the time-scale of the meta controller that proved to be more efficient and flexible than alternative approaches for interrupting subtasks and for dealing with delayed sequential decision problems. Our empirical results on a stochastic 3D domain showed that our architecture generalizes well to longer sequences of instructions as well as unseen instructions. Although our hierarchical RL architecture was demonstrated in the simple setting where the set of instructions should be executed sequentially, we believe that our key ideas are not limited to this setting but can be extended to richer forms of instructions.

\section*{Acknowledgement} This work was supported by NSF grant IIS-1526059. Any opinions, findings, conclusions, or recommendations expressed here are those of the authors and do not necessarily reflect the views of the sponsor.

\bibliography{references}
\bibliographystyle{abbrvnat}
\clearpage
\appendix
\onecolumn

\newcommand{\ck}{\checkmark}
\section{Experimental Setting for Parameterized Tasks} \label{sec:injecting-prior}
The episode terminates after 50 steps for `Independent' and `Object-dependent' cases and 70 steps for `Inter/Extrapolation' case. The agent receives a positive reward ($+1$) when it successfully finishes the given task and receives a time penalty of $-0.1$ for each step. The details of each generalization scenario are described below.

\paragraph{Independent.}
The semantics of the tasks are consistent across all types of target objects. The training set of tasks is shown in Table~\ref{tab:task-independent}. Examples of analogies used in this experiment are:
\vspace{-8pt}
\begin{itemize}[leftmargin=*]
\setlength\itemsep{0em}
\item {[Visit, X] : [Visit, Y] :: [Transform, X] : [Transform, Y]}
\item {[Visit, X] : [Visit, Y] $\neq$ [Transform, X] : [Pick up, Y]}
\item {[Visit, X] $\neq$ [Visit, Y]},
\vspace{-4pt}
\end{itemize}
where X and Y can be any object type (X $\neq$ Y).

\paragraph{Object-dependent.}
We divided objects into two groups: Group A and Group B. Given `Interact with' action, Group A should be picked up, whereas Group B should be transformed by the agent. The training set of tasks with groups for each object is shown in Table~\ref{tab:task-dependent}. Examples of analogies used in this experiment are:
\vspace{-4pt}
\begin{itemize}[leftmargin=*]
\setlength\itemsep{0em}
\item {[Visit, Sheep] : [Visit, Cat] :: [Interact with, Sheep] : [Interact with, Cat]}
\item {[Visit, Sheep] : [Visit, Greenbot] $\neq$ [Interact with, Sheep] : [Interact with, Greenbot]} 
\item {[Visit, Horse] : [Visit, Greenbot] :: [Interact with, Horse] : [Interact with, Greenbot]} 
\vspace{-4pt}
\end{itemize}
Note that the second example implies that Sheep and Greenbot should be treated in a different way because they belong to different groups. Given such analogies, the agent can learn to interact with Cat as it interacts with Sheep and interact with Greenbot as it interacts with Horse.

\paragraph{Inter/Extrapolation.}
In this experiment, a task is defined by three parameters: action, object, and number. The agent should repeat the same subtask for a given number of times. The agent is trained on all configurations of actions and target objects. However, only a subset of numbers is used during training. In order to interpolate and extrapolate, we define analogies based on simple arithmetic such as:
\vspace{-4pt}
\begin{itemize}[leftmargin=*]
\setlength\itemsep{0em}
\item {[Pick up, X, 1] : [Pick up, X, 6] :: [Pick up, X, 2] : [Pick up, X, 7]}
\item {[Pick up, X, 3] : [Pick up, X, 6] :: [Transform, Y, 4] : [Transform, Y, 7]}
\item {[Pick up, X, 1] : [Pick up, X, 3] $\neq$ [Transform, Y, 2] : [Transform, Y, 3]}
\vspace{-4pt}
\end{itemize}

\begin{table}[H]
\centering
\begin{minipage}[t]{0.42\linewidth}
\centering
\small
\begin{tabular}[t]{r||c|c|c}
\hline
		& Visit	& Pick up & Transform \\ \hline
Sheep	& \ck & 		& \ck \\ 
Horse	& \ck & 		& \ck \\ 
Pig		& \ck & \ck 	& 	  \\ 
Box		& \ck & \ck 	& 	  \\ 
Cat		& 	  & \ck 	& \ck \\ 
Greenbot	& 	  & \ck 	& \ck \\ \hline
\end{tabular}
\vspace{-5pt}
\caption{Training set of tasks for the `Independent' case. }
\label{tab:task-independent}
\end{minipage}
\hspace{10pt}
\begin{minipage}[t]{0.55\linewidth}
\centering
\small
\begin{tabular}[t]{r||c|c|c|c}
\hline
		    & Visit	& Pick up & Transform & Interact with \\ \hline
Sheep (A)   & \ck & \ck	& \ck & \ck \\ 
Horse (B)	& \ck & \ck	& \ck & \ck \\ 
Pig	(A)	    & \ck & \ck & \ck &	\ck \\ 
Box	(B)	    & \ck & \ck & \ck &	\ck \\ 
Cat	(A)	    & \ck & \ck & \ck \\ 
Greenbot (B)	& \ck & \ck & \ck \\ \hline
\end{tabular}
\vspace{-5pt}
\caption{Training set of tasks for the `Object-dependent' case. }
\label{tab:task-dependent}
\hfill 
\end{minipage}
\end{table}

\clearpage
\section{Experimental Setting for Instruction Execution} \label{sec:environment-task}
The episode terminates after 80 steps during training of the meta controller. A box appears with probability of 0.03 and disappears after 20 steps. During evaluation on longer instructions, the episode terminates after 600 steps.
We constructed a training set of instructions and an unseen set of instructions as described in Table~\ref{tab:instruction-split}. We generated different sequences of instructions for training and evaluation by sampling instructions from such sets of instructions.

\begin{table}[H]
\centering
\small
\begin{tabular}{r||c|c|c|c|c|c|c}
\hline
		& Visit	& Pick up & Pick up 2 & Pick up 3 & Transform & Transform 2 & Transform 3 \\ \hline
Sheep	& \ck & 		& & & \ck & \ck & \ck \\ 
Horse	& \ck & 		& & & \ck & \ck & \ck \\ 
Pig		& \ck & \ck 	& \ck & \ck & & & 	  \\ 
Cat		& 	  & \ck & \ck & \ck 	& \ck & \ck & \ck \\ 
Greenbot	& 	  & \ck & \ck & \ck 	& \ck & \ck & \ck \\ \hline
\end{tabular}
\vspace{-5pt}
\caption{Training set of instructions. }
\label{tab:instruction-split}
\vspace{-5pt}
\end{table}

\section{Experiment on 2D Grid-World} \label{sec:environment-task}
\begin{wrapfigure}{r}{0.5\linewidth}
 	\centering
 	\vspace{-5pt}
 	\includegraphics[width=\linewidth]{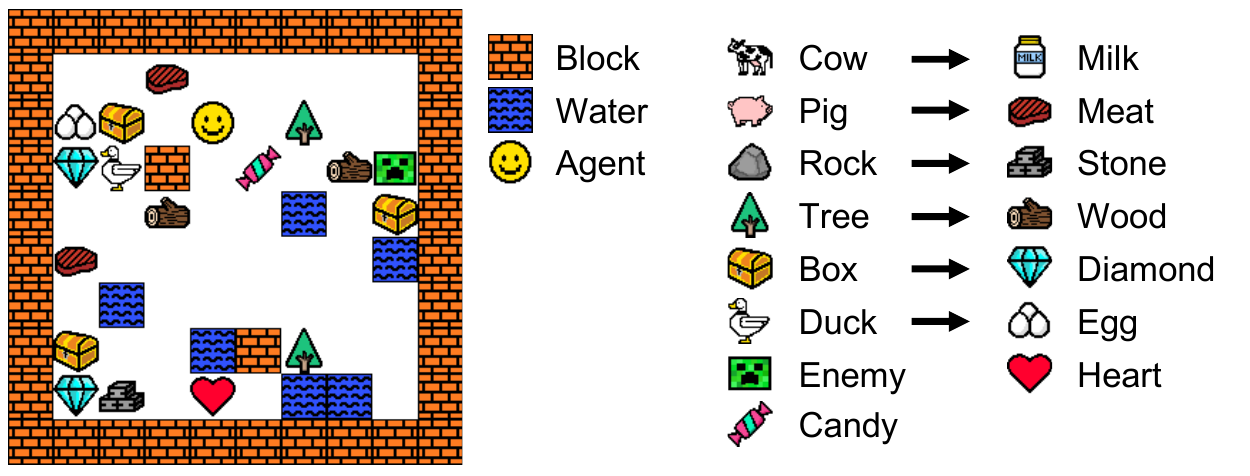} 
 	\caption{Example of 2D grid-world with object specification. The arrows represent the outcome of object transformation. Objects without arrows disappear when transformed. The agent is not allowed to go through blocks and gets a penalty for going through water. } 
 	\label{fig:environment}
 	\vspace{-15pt}
\end{wrapfigure}
\paragraph{Environment.} 
To see how our approach can be generally applied to different domains, we developed a 2D grid-world based on MazeBase~\citep{sukhbaatar2015mazebase} where the agent can interact with many objects, as illustrated in Figure~\ref{fig:environment}. Unlike the original MazeBase, an observation is represented as a binary 3D tensor: $\textbf{x}_t\in \mathbb{R}^{18 \times 10 \times 10}$ where $18$ is the number of object types and $10 \times 10$ is the size of the grid world. Each channel is a binary mask indicating the presence of each object type. There are agent, blocks, water, and 15 types of objects with which the agent can interact, and all of them are randomly placed for each episode. 

The agent has 13 primitive actions: \emph{No-operation}, \emph{Move} (North/South/West/East, referred to as ``NSWE"), \emph{Pick up} (NSWE), and \emph{Transform} (NSWE). \emph{Move} actions move the agent by one cell in the specified direction. 
\emph{Pick up} actions remove the adjacent object in the corresponding relative position, and \emph{Transform} actions either remove it or transform it to another object depending on the object type, as shown in Figure~\ref{fig:environment}. 

The agent receives a time penalty ($-0.1$) for each time-step. Water cells act as obstacles which give $-0.3$ when the agent visits them. The agent receives $+1$ reward when it finishes all instructions in the correct order. Throughout the episode, an enemy randomly appears, moves, and disappears after 10 steps. Transforming an enemy gives $+0.9$ reward.  

\paragraph{Evaluation on Parameterized Tasks.}
As in the main experiment, we evaluated the parameterized skill on seen and unseen parameterized tasks separately using the same generalization scenarios. As shown in Table~\ref{tab:2d-subtask}, the parameterized skill with analogy-making can successfully generalize to unseen tasks both in independent and object-dependent scenarios.

\begin{table}[H]
\centering
\begin{minipage}{0.7\textwidth}
\centering
\setlength{\tabcolsep}{3pt}
\begin{tabular}{c|c||c|c}
  \hline
Scenario & Analogy & Train & Unseen \\ \hline 
\multirow{2}{*}{Independent} & $\times$ & \textbf{0.56} $(99.9\%)$ & -1.88 $(49.6\%)$ \\
& \checkmark & \textbf{0.56} $(99.9\%)$ & \textbf{0.55} $(99.6\% )$ \\ \hline
\multirow{2}{*}{Object-dependent} & $\times$ & \textbf{0.55} $(99.9\%)$ & -3.23 $(43.2\%)$ \\
& \checkmark & \textbf{0.55} $(99.9\%)$ & \textbf{0.55} $(99.5\%)$ \\ \hline
\end{tabular}
  \vspace{-5pt}
  \caption{Performance on parameterized tasks. Each entry shows `Average reward (Success rate)'. We assume an episode is successful only if the agent successfully finishes the task and its termination predictions are correct throughout the whole episode. }
  \label{tab:2d-subtask}
  \vspace{-10pt}
\end{minipage}
\end{table}

\begin{figure}[H]
    	\centering
    	\begin{subfigure}{0.18\textwidth}
			\centering		    
		    \includegraphics[width=\textwidth]{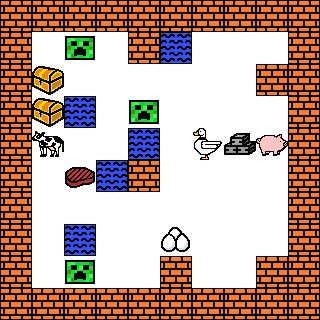} 
		    \caption{Observation}
  		\end{subfigure}
  		\begin{subfigure}{0.18\textwidth}
  			\centering
		    \includegraphics[width=\textwidth]{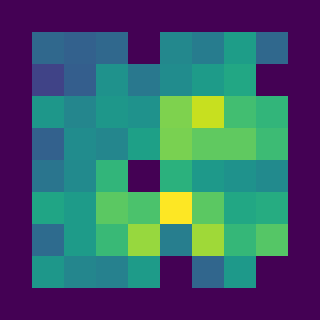} 
		    \caption{Visit egg}
  		\end{subfigure}
  		\begin{subfigure}{0.18\textwidth}
  			\centering
		    \includegraphics[width=\textwidth]{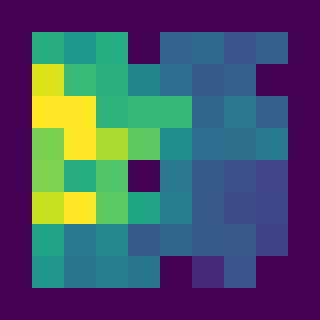} 
		    \caption{Pick up cow}
  		\end{subfigure}
  		\begin{subfigure}{0.18\textwidth}
  			\centering
		    \includegraphics[width=\textwidth]{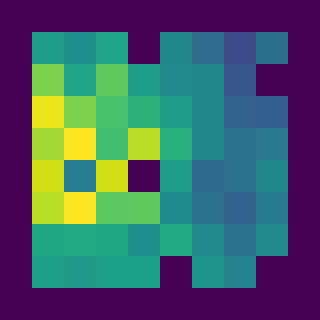} 
		    \caption{Transform meat}
  		\end{subfigure}
  		\vspace{-5pt}
		\caption{Value function visualization given unseen tasks. (b-d) visualizes learned values for each position of the agent in a grid world (a). The agent estimates high values around the target object in the world.} 
		\label{fig:value}
		\vspace{-5pt}
\end{figure}

We visualized the value function learned by the critic network of the parameterized skill in Figure~\ref{fig:value}. As expected from its generalization performance, our parameterized skill trained with analogy-making objective learned high values around the target objects given unseen tasks.

\paragraph{Evaluation on Instruction Execution.}
There are 5 types of instructions: Visit X, Pick up X, Transform X, Pick up all X, and Transform all X, where `X' is the target object type. While the first three instructions require the agent to perform the corresponding subtask, the last two instructions require the agent to repeat the same subtask until the target objects completely disappear from the world. 

The overall result is consistent with the result on 3D environment as shown in Figure~\ref{fig:curve} and Table~\ref{tab:performance}. The flat baseline learned a sub-optimal policy that transforms or picks up all target objects in the world even if there is no `all' adverb in the instructions. This sub-optimal policy unnecessarily removes objects that can be potentially target objects in the future instructions. This is why the performance of the flat baseline drastically decreases as the number of instructions increases in Figure~\ref{fig:curve}. 
Our architecture with learned time-scale (`Hierarchical-Dynamic') performs much better than `Hierarchical-Short' baseline which updates the subtask at every time-step. This also suggests that learning to operate in a large-time scale is crucial for dealing with delayed reward. 
We also observed that our agent learned to deal with enemies whenever they appear, and thus it outperforms the `Shortest Path' method which is near-optimal in executing instructions while ignoring enemies. 

\begin{figure}[H]
    \centering
   	\hspace{-5pt}
    \begin{subfigure}{0.20\textwidth}
    	\centering
	    \includegraphics[width=\textwidth]{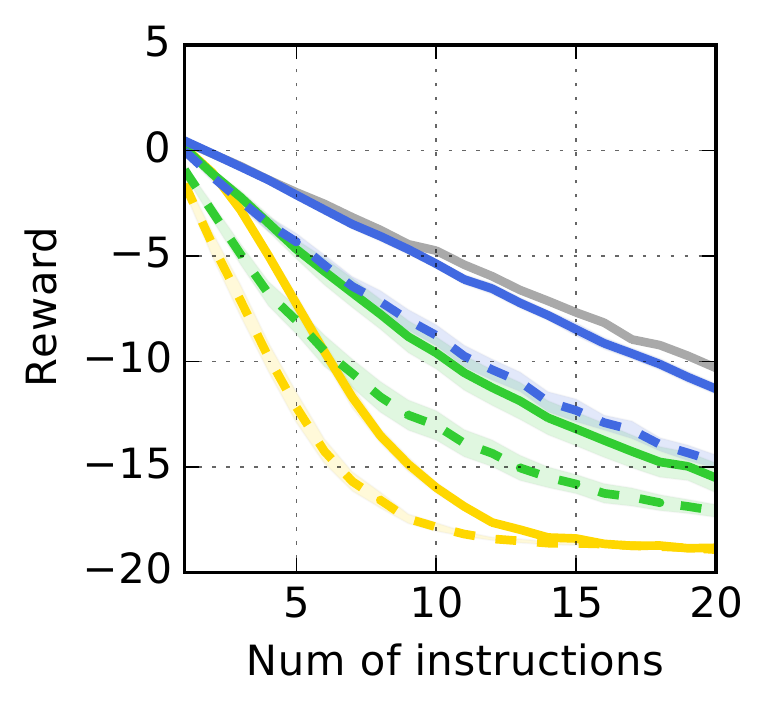} 
   	\end{subfigure}
   	\hspace{-7pt}
	\begin{subfigure}{0.20\textwidth}
		\centering
	    \includegraphics[width=\textwidth]{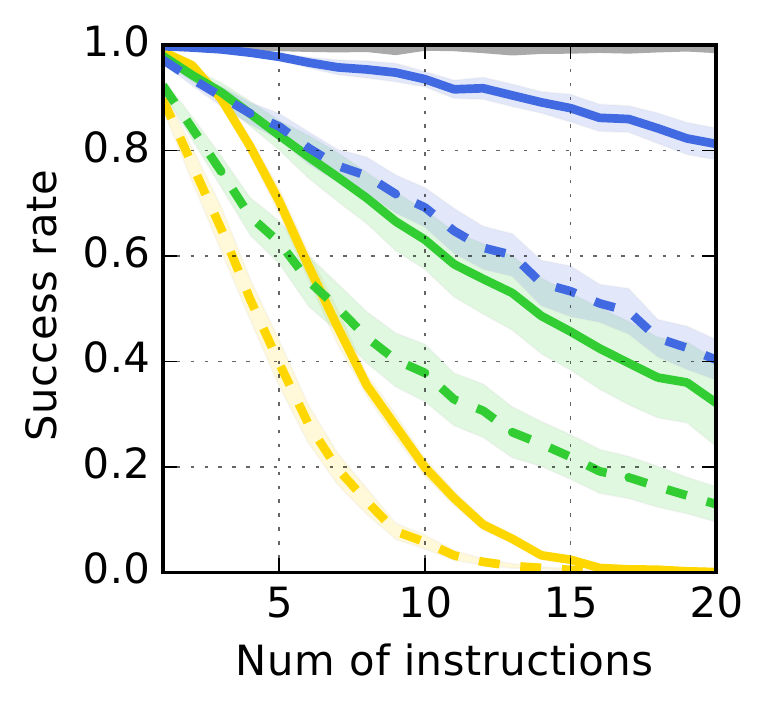} 
	\end{subfigure}
   	\hspace{-7pt}
	\begin{subfigure}{0.20\textwidth}
		\centering
	    \includegraphics[width=\textwidth]{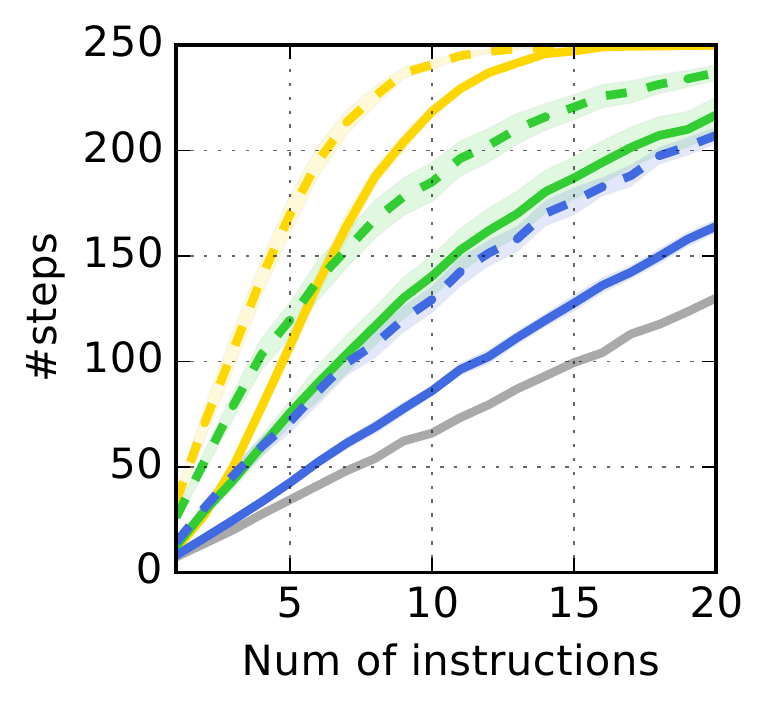} 
	\end{subfigure}
   	\hspace{-7pt}
	\begin{subfigure}{0.20\textwidth}
		\centering
	    \includegraphics[width=\textwidth]{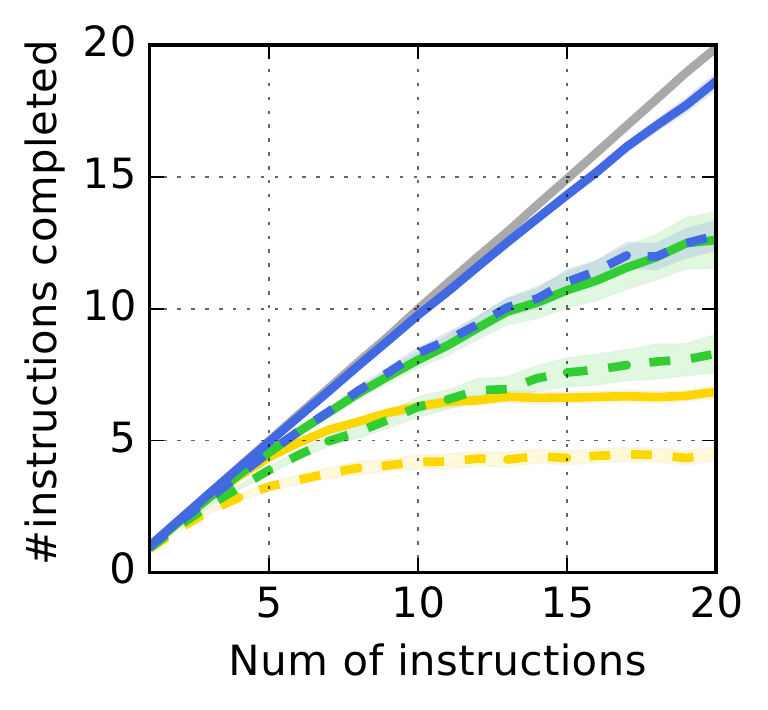} 
	\end{subfigure}
   	\hspace{-5pt}
	\begin{subfigure}{0.20\textwidth}
		\centering
	    \raisebox{4mm}{\includegraphics[width=\textwidth]{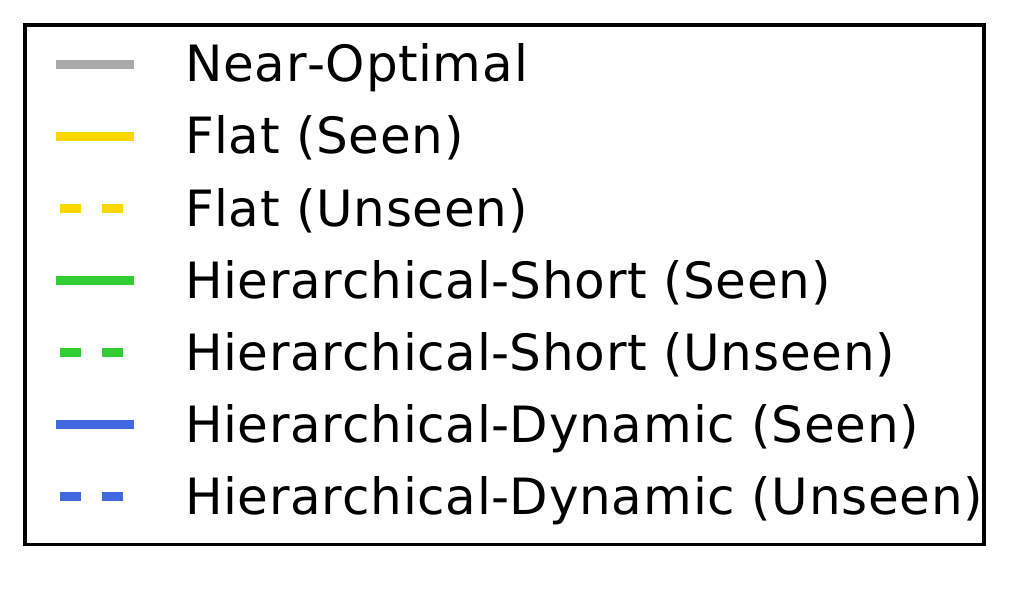}}
	\end{subfigure}
	\vspace{-5pt}
	\caption{Performance per number of instructions.}
	\vspace{-15pt}
	\label{fig:curve}
\end{figure}

\begin{table}[H]
\centering
\setlength{\tabcolsep}{3pt}
\begin{tabular}{r||c|cc}
\hline
& Train & Test (Seen) & Test (Unseen) \\ 
\hline
 \#Instructions & 4 & 20 & 20 \\ \hline
 Shortest Path & -1.62 $(99.7\%)$ & \multicolumn{2}{c}{-11.94 $(99.4\%)$} \\ 
 Near-Optimal & -1.34 $(99.5\%)$ & \multicolumn{2}{c}{-10.30 $(99.3\%)$} \\ \hline

 Flat & -2.38 $(76.0\%)$ & -18.83 $(0.1\%)$ & -18.92 $(0.0\%)$ \\ 
 Hierarchical-Short & -1.74 $(81.0\%)$ & -15.89 $(28.0\%)$ & -17.23 $(11.3\%)$ \\ 
 Hierarchical-Dynamic & \textbf{-1.26} $(95.5\%)$ & \textbf{-11.30} $(81.3\%)$ & \textbf{-14.75} $(40.3\%)$ \\ 
\hline 
\end{tabular}
\vspace{-5pt}
\caption{Performance of meta controller. Each entry in the table represents reward with success rate in parentheses averaged over 10-best runs among 20 independent runs. `Shortest Path' is a hand-designed policy which executes instructions optimally based on the shortest path but ignores enemies. `Near-Optimal' is a near-optimal policy that executes instructions based the shortest path and transforms enemies when they are close to the agent.  }
\label{tab:performance}
\vspace*{-0.15in}
\end{table}

\clearpage
\section{Details of Learning Objectives} \label{sec:detail-learning}
\subsection{Parameterized Skill} \label{sec:multi-training}
The parameterized skill is first trained through \textit{policy distillation}~\citep{rusu2015policy,parisotto2015actor} and fine-tuned using actor-critic method~\citep{konda1999actor} with generalized advantage estimation (GAE)~\citep{schulman2015high}. The parameterized skill is also trained to predict whether the current state is terminal or not through binary classification objective. 

The idea of policy distillation is to first train separate teacher policies ($\pi_T^{g}(a|s)$) for each task ($g$) through reinforcement learning and train a multi-task policy ($\pi^{g}_{\phi}(a|s)$) to mimic teachers' behavior by minimizing KL divergence between them as follows:
\begin{equation}
\nabla_{\phi} \mathcal{L}_{RL} = \mathbb{E}_{g \sim \mathcal{U}}\left[\mathbb{E}_{s \sim \pi^{g}_{\phi}}\left[\nabla_{\phi} D_{KL}\left(\pi_T^{g} || \pi^{g}_{\phi} \right) + \alpha \nabla_{\phi} \mathcal{L}_{term} \right] \right],
\end{equation} 
where $D_{KL}\left(\pi_T^{g} || \pi^{g}_{\phi}\right) = \sum_{a} \pi_T^{g}(a|s) \log \frac{\pi_T^{g}(a|s)}{\pi^{g}_{\phi} (a|s)}$ and $\mathcal{U}\subset \mathcal{G}$ is the training set of tasks. $\mathcal{L}_{term} = -\log \beta_{\phi}\left(s_t,g \right)=-\log P_\phi\left(s_t \in \mathcal{T}_g \right)$ is the cross-entropy loss for termination prediction. Intuitively, we sample a mini-batch of tasks ($g$), use the parameterized skill to generate episodes, and train it to predict teachers' actions. This method has been shown to be efficient for multi-task learning. 

After policy distillation, the parameterized skill is fine-tuned through actor-critic with generalized advantage estimation (GAE)~\citep{schulman2015high} as follows:
\begin{equation}
\nabla_{\phi} \mathcal{L}_{RL} = 
\mathbb{E}_{g \sim \mathcal{U}}\left[\mathbb{E}_{s \sim \pi^{g}_{\phi}} \left[ -\nabla_{\phi}\log\pi_{\phi}\left(a_t|s_t,g\right)\hat{A}^{(\gamma,\lambda)}_t  + \alpha \nabla_{\phi} \mathcal{L}_{term} \right] \right],
\label{eq:multi-gradient}
\end{equation} 
where $\hat{A}^{(\gamma,\lambda)}_t=\sum^{\infty}_{l=0}(\gamma \lambda)^l \delta^{V}_{t+l}$ and $ \delta^{V}_t=r_t + \gamma V^{\pi}(s_{t+1};\phi ')-V^{\pi}(s_t; \phi ')$.  $\phi '$ is optimized to minimize $\mathbb{E}\left[ \left(R_t - V^{\pi}(s_t;\phi ') \right)^2 \right]$. $\gamma,\lambda \in \left[0, 1\right]$ are a discount factor and a weight for balancing between bias and variance of the advantage estimation. 

The final update rule for the parameterized skill is:
\begin{align} 
\Delta \phi  \propto -\left(\nabla_{\phi} \mathcal{L}_{RL} + \xi \nabla_{\phi} \mathcal{L}_{AM}\right), \label{eq:multi-obj}
\end{align} 
where $\mathcal{L}_{AM} = \mathcal{L}_{sim} + \rho_1 \mathcal{L}_{dis}  + \rho_2 \mathcal{L}_{diff}$ is the analogy-making regularizer defined as the weighted sum of three objectives described in the main text.  $\rho_1,\rho_2,\xi$ are hyperparameters for each objective.

\subsection{Meta Controller} \label{sec:meta-training}
Actor-critic method with GAE is used to update the parameter of the meta controller as follows:
\begin{align}
\nabla_{\theta}\mathcal{L}_{RL} = -
\begin{cases}
\mathbb{E}\left[ c_t\left(\sum_{i}\nabla_{\theta}\log\pi_{\theta}\left(g^{(i)}_t|\textbf{h}_t,\textbf{r}_t\right) + \nabla_{\theta}\log P\left(\textbf{l}_t|\textbf{h}_{t}\right) \right)\hat{A}^{(\gamma,\lambda)}_t 
\right.  & \multirow{2}{*}{\mbox{(Hard)}} \\ \left. \vphantom{\left[\sum^{n}_i\right]} 
+ \nabla_{\theta}\log P \left(c_t|\textbf{s}_t,\textbf{h}_{t-1}\right)\hat{A}^{(\gamma,\lambda)}_t + \eta \nabla_{\theta}\left\Vert \sigma \left( \varphi^{update}\left(\textbf{s}_t,\textbf{h}_{t-1} \right) \right) \right\Vert_1 \right]   \\
 \mathbb{E} \left[\sum_{i}\nabla_{\theta}\log\left[c_t\pi_{\theta}\left(g^{(i)}_t|\tilde{\textbf{h}}_t,\tilde{\textbf{r}}_t\right)+(1-c_t)g^{(i)}_{t-1}\right] \hat{A}^{(\gamma,\lambda)}_t \right] & \mbox{(Soft)},
\end{cases}
\label{eq:meta-rl}
\end{align}
where $c_t\sim P\left(c_t\vert \textbf{s}_t,\textbf{h}_{t-1}\right) \propto \sigma \left( \varphi^{update}\left(\textbf{s}_t,\textbf{h}_{t-1} \right) \right) $, and $P\left(\textbf{l}_t|\textbf{h}_t\right) \propto \mbox{Softmax}\left( \varphi^{shift}(\textbf{h}_t) \right)$.  We applied L1-penalty to the probability of update to penalize too frequent updates, and $\eta$ is a weight for the update penalty. 

The final update rule for the meta controller is:
\begin{align} 
\Delta \theta  \propto -\left(\nabla_{\theta} \mathcal{L}_{RL} + \xi \nabla_{\theta} \mathcal{L}_{AM}\right), \label{eq:multi-obj}
\end{align} 
where $\mathcal{L}_{AM}$ is the analogy-making objective.

\clearpage
\section{Architectures and Hyperparameters} \label{sec:arch-hyper}

\paragraph{Background: Multiplicative Interaction}
For combining condition variables into a neural network (e.g., combining task embedding into the convolutional network in the parameterized skill), we used a form of multiplicative interaction instead of concatenating such variables as suggested by~\cite{memisevic2010learning,oh2015action}. This is also related to \textit{parameter prediction} approaches where the parameters of the neural network is produced by condition variables (e.g., exempler, class embedding). This approach has been shown to be effective for achieving zero-shot and one-shot generalization in image classification problems~\citep{lei2015predicting,bertinetto2016learning}.
More formally, given an input ($\textbf{x}$), the output ($\textbf{y}$) of a convolution and a fully-connected layer with parameters predicted by a condition variable ($\textbf{g}$) can be written as:
\begin{align*} 
\mbox{Convolution: } & \textbf{y} = \varphi\left(\textbf{g}\right) \ast \textbf{x} + \textbf{b} \\
\mbox{Fully-connected: } & \textbf{y} = \textbf{W}'\mbox{diag}\left( \varphi\left(\textbf{g}\right) \right)\textbf{W}  \textbf{x} + \textbf{b},
\end{align*}
where $\varphi$ is the embedding of the condition variable learned by a multi-layer perceptron (MLP). Note that we use matrix factorization (similar to \citep{memisevic2010learning}) to reduce the number of parameters for the fully-connected layer. Intuitively, the condition variable is converted to the weight of the convolution or fully-connected layer through multiplicative interactions. We used this approach both in the parameterized skill and the meta controller. The details are described below.

\paragraph{Parameterized skill.} The teacher architecture used for policy distillation is Conv1(32x8x8-4)-Conv2(64x5x5-2)-LSTM(64).\footnote{For convolution layers, NxKxK-S represents N kernels with size of KxK and stride of S. The number in LSTM represents the number of hidden units.} 
The network has two fully-connected output layers for actions and value (baseline) respectively. The parameterized skill architecture consists of Conv1(16x8x8-4)-Conv2(32x1x1-1)-Conv3(32x1x1-1)-Conv4(32x5x5-2)-LSTM(64). The parameterized skill takes two task parameters ($\textbf{g}=\left[g^{(1)},g^{(2)}\right]$) as additional input and computes $\varphi(\textbf{g})=\mbox{ReLU}(\textbf{W}^{(1)}g^{(1)}\odot\textbf{W}^{(2)}g^{(2)})$ to compute the subtask embedding. It is further linearly transformed into the weights of Conv3 and the (factorized) weight of LSTM through multiplicative interaction as described above. Finally, the network has three fully-connected output layers for actions, termination probability, and baseline, respectively.

We used RMSProp optimizer with the smoothing parameter of $0.97$ and epsilon of $1e-6$. When training the teacher policy through actor-critic, we used a learning rate of $2.5e-4$. For training the parameterized skill, we used a learning rate of $2.5e-4$ and $1e-4$ for policy distillation and actor-critic fine-tuning respectively. We used $\tau_{dis}=\tau_{diff}=3,\alpha=0.1$ for analogy-making regularization and the termination prediction objective respectively. $\gamma=0.99$ and $\lambda=0.96$ are used as a discount factor and a balancing weight for GAE. 16 threads with batch size of 8 are used to run $16\times8$ episodes in parallel, and the parameter is updated after each run (1 iteration = $16\times8$ episodes). For better exploration, we applied entropy regularization with a weight of $0.1$ and linearly decreased it to zero for the first 7500 iterations. The total number of iterations was 15,000 for both policy distillation and actor-critic fine-tuning.

\paragraph{Meta Controller.}
The meta controller consists of Conv1(16x8x8-4)-Conv2(32x1x1-1)-Conv3(32x1x1-1)-Pool(5)-LSTM(256). The embedding of previously selected subtask ($\varphi(\textbf{g}_{t-1})$), the previously retrieved instruction ($\textbf{r}_{t-1}$), and the subtask termination ($b_t$) are concatenated and given as input for one-layer MLP to compute a 256-dimensional joint embedding. This is further linearly transformed into the weights of Conv3 and LSTM through multiplicative interaction. The output of FC1 is used as the context vector ($\textbf{h}_t$). We used the bag-of-words (BoW) representation as a sentence embedding which computes the sum of all word embeddings in a sentence: $\varphi^{w}\left(\textbf{m}_i\right)=\sum^{|\textbf{m}_i|}_{j=1}\textbf{W}^{m}w_j$ where $\textbf{W}^{m}$ is the word embedding matrix, each of which is 256-dimensional. An MLP with one hidden layer with 256 units is for $\varphi^{shift}$, a fully-connected layer is used for $\varphi^{update}$. $\varphi^{goal}$ is an MLP with one hidden layer with 256 units that takes the concatenation of $\textbf{r}_t$ and $\textbf{h}_t$ as an input and computes the probabilities over subtask parameters as the outputs. The baseline network is a linear regression from the concatenation of the memory pointer $\textbf{p}_t$, a binary mask indicating the presence of given instruction, and the final hidden layer (256 hidden units in $\varphi^{goal}$).

We used the same hyperparameters used in the parameterized skill except that the batch size was 32 (1 iteration = $16\times32$ episodes). We trained the soft-architecture (with soft-update) with a learning rate of $2.5e-4$ using curriculum learning for 15,000 iterations and a weight of $0.015$ for entropy regularization, and fine-tuned it with a learning rate of $1e-4$ without curriculum learning for 5,000 iterations. Finally, we initialized the hard-architecture (with hard-update) to the soft-architecture and fine-tuned it using a learning rate of $1e-4$ for 5,000 iterations. $\eta=0.001$ is used to penalize frequent update decision in Eq~\eqref{eq:meta-rl}. 

\paragraph{Flat Controller.}
The flat controller architecture consists of the same layers used in the meta controller with the following differences. The previously retrieved instruction ($\textbf{r}_{t-1}$) is transformed through an MLP with two hidden layers to compute the weight of Conv3 and LSTM. The output is probabilities over primitive actions.

\paragraph{Curriculum Learning.} For training all architectures, we randomly sampled the size of the world from $\{5,6,7,8\}$, the density of walls are sampled from $\left[0,0.1\right]$, and the density of objects are sampled from $\left[0.1,0.8\right]$ during training of parameterized skill and $\left[0,0.15\right]$ during training of the meta controller. We sampled the number of instructions from $\{1,2,3,4\}$ for training the meta controller. The sampling range was determined based on the success rate of the agent. 
\end{document}